\documentclass{ieeeaccess}
\usepackage{cite}
\usepackage{amsmath,amssymb,amsfonts}
\usepackage{algorithmic}
\usepackage{graphicx}
\usepackage{textcomp}
\usepackage{algorithmic}
\usepackage{subfigure}
\usepackage{graphicx}
\usepackage{amssymb}
\usepackage{url}
\usepackage{enumerate}
\usepackage[normalem]{ulem}

\usepackage{rotating}

\def\BibTeX{{\rm B\kern-.05em{\sc i\kern-.025em b}\kern-.08em
    T\kern-.1667em\lower.7ex\hbox{E}\kern-.125emX}}
\begin{document}
\history{Date of publication xxxx 00, 0000, date of current version xxxx 00, 0000.}
\doi{10.1109/ACCESS.2017.DOI}

\title{Survey of State-of-the-Art Mixed Data Clustering Algorithms}
\author{\uppercase{AMIR AHMAD}\authorrefmark{1} and \uppercase{SHEHROZ KHAN} \authorrefmark{2},}
\address[1]{College of Information Technology, United Arab Emirates University,Al-Ain, UAE  (e-mail: amirahmad@uaeu.ac.ae)}
\address[2]{Toronto Rehabilitation Institute, University Health Network, 550, University Avenue, Toronto, Canada, (e-mail: shehroz.khan@uhn.ca)}
\tfootnote{``This work was supported by a UAE university Start-up grant (grant number G00002668; fund number 31T101.''}

\markboth
{Author \headeretal: Preparation of Papers for IEEE TRANSACTIONS and JOURNALS}
{Author \headeretal: Preparation of Papers for IEEE TRANSACTIONS and JOURNALS}

\corresp{Corresponding author: AMIR AHMAD (e-mail: amirahmad@uaeu.ac.ae).}

\begin{abstract}
Mixed data comprises both numeric and categorical features, and mixed datasets  occur frequently in many domains, such as health, finance, and marketing. Clustering is often applied to mixed datasets to find structures and to group similar objects for further analysis. However, clustering mixed data is challenging because it is difficult to directly apply mathematical operations, such as summation or averaging, to the feature values of these datasets. In this paper, we present a taxonomy for the study of mixed data clustering algorithms by identifying five major research themes. We then present a state-of-the-art review of the research works within each research theme. We analyze the strengths and weaknesses of these methods with pointers for future research directions. Lastly, we present an in-depth analysis of the overall challenges in this field, highlight open research questions and discuss guidelines to make progress in the field.
\end{abstract}


\titlepgskip=-15pt

\maketitle

\section{Introduction}
Clustering is an unsupervised machine learning technique used to group unlabeled data into clusters that contain data points that are `\textit{similar}' to each other and `\textit{dissimilar}' from those in other clusters \cite{kmeanclusterinitialization1,ClusteringbookJain}. Many clustering algorithms can only handle data that contain either numeric or categorical feature values \cite{Booksmachine,WEKAwitten1}. Numeric features  can take real values, such as height, weight, and distance. Categorical features represent data that can be divided into a fixed number of categories, such as color, race, sex, profession, and blood group. 
Clustering algorithms group data points into clusters using some notion of `\textit{similarity}', which can be as simple as the Euclidean distance. To compute the similarity between numeric feature values, mathematical operations (such as distances, angles, summation, or mean) are applied to them. Distance-based similarity measures are mostly used for numeric data points. Generally, categorical feature values are not inherently ordered (for example, the categorical values, red and blue).  It is not possible to directly compute the distance between two categorical feature values. Therefore, computing distance-based similarity measures for categorical data is a challenging task \cite{Categoricalsimilarity}. Nevertheless, several methods have been suggested in the literature for computing the similarity of data points containing categorical features \cite{Categoricalsimilarity}.

Many real-world datasets contain both numeric and categorical features; they  are called \textit{mixed datasets}. 
Mixed data occur frequently in many applications, such as health, marketing, medical, and finance \cite{AmirLipika2007, SocioMixed,Morlini2010}. Therefore, developing machine learning algorithms that  can handle such data has become important. Clustering is a natural choice for practitioners to determine groups of mixed data points for further data analysis. However, the problem of computing the similarity of two data points becomes more difficult when the dataset contains both numerical and categorical features. An example snapshot of a typical mixed dataset is shown in Table \ref{tab:toy}. This sample dataset has four features-; \textit{Height} and \textit{Weight} are numeric features, whereas \textit{Blood Group} and \textit{Profession} are categorical features. A simple strategy to find similarity between two data points in this dataset is to split the numeric and categorical parts and find the Euclidean distance between two data points for the numeric features and the Hamming distance for the categorical features \cite{Huang1998}. This will enable one to find the similarity between numeric and categorical feature values, albeit separately. The next step is to combine these two measures to get one value that represents the distance between two mixed data points.
However, combining these two types of distances directly is non-trivial, because it is not clear, 
\begin{enumerate}[(i)]
    \item whether both of the distance measures calculate a `similar' type of similarity, or
    \item whether the scales of these distances are similar. Therefore, the proportions in which the two distance measures are combined is non-obvious. 
\end{enumerate}
Hence, as the notion of similarity is not clearly defined for mixed data, performing clustering on them remains challenging. 

\begin{table}[!ht]
\centering
\caption{An example mixed dataset.}
\label{tab:toy}
\begin{tabular}{|l|l|l|l|}
\hline
Weight &Height  & Blood  & Profession
\\ (kg) & (m) & Group   & 
\\
\hline
80.6 & 1.85 & B+ & Teaching
\\73.6&1.72 & A+ & Teaching
\\70.8 & 1.79 & B+ & Medical
\\85.9 & 1.91& A-& Sportsman
\\83.4& 1.65 & A+& Medical
\\
\hline
\end{tabular}
\end{table}

Two major focuses of most mixed data clustering algorithms are (i) to find innovative ways to define novel measures of similarity between mixed features, and (ii) to perform  clustering using existing or new techniques. Some of the earliest techniques of mixed clustering were direct extensions of partitional clustering algorithms (for example, K-means) \cite{Huang1997,Huang1998}. Since then, many new research themes have evolved and developed in this field of research. 
In this paper, we present a taxonomy to identify five broad research themes for mixed data clustering algorithms based on the methodology used to cluster mixed datasets. Using this taxonomy, we present a comprehensive review of clustering algorithms within each research theme. We present a critical analysis of the different types of mixed data clustering algorithms, and discuss their functions, strengths and weaknesses.
We further identify challenges and open research questions among the different types of mixed data clustering algorithms and discuss on opportunities to make advances in the field. 
The main contributions of our paper are as follows:
\begin{itemize}
    \item We identify a few other survey papers on mixed data clustering and differentiate them in our comprehensive literature review in terms of scope, taxonomy, research areas, applications, and vision for future work
    \item We present a new taxonomy to identify five broad research themes for the study of mixed data clustering and present a critical review of the literature on these research themes.
    \item We present a detailed analysis of the application areas in which mixed data clustering may have major impact.
    \item We present an in-depth analysis of ensuing challenges, open research questions and guidelines to be adopted to make progress in the field.
\end{itemize}

\section{Survey of other review papers}
Few review articles on mixed data clustering have been published recently. However, they are not detailed and they concentrate on specific types of clustering algorithms. Velden \textit{et al.} \cite{Reviewdistance} study five distance-based clustering algorithms for mixed data on three mixed datasets. They conclude that there is no single clustering approach that performs well for all the datasets. The review presented by Fuss \textit{et al.} \cite{Reviewmodelbased} concentrate only on partitional clustering and model-based clustering for mixed datasets.  Balaji and Lavanya present a short review paper on mixed data clustering \cite{mixedddatareview}. Many important mixed data clustering algorithms and research themes are omitted from in the paper. The paper also does not discuss the challenges or the future directions in this area. The review paper by Miyamoto \textit{et al.} \cite{miyamotoreview}  discusses only the basic concepts of clustering, no mixed data clustering algorithm is described in the paper. The published literature review on mixed data clustering show several drawbacks:
\begin{itemize}
    \item Most of these papers fail to identify concrete research themes or taxonomy to pave the way for performing systematic research in the field.
    \item None of these papers are comprehensive in their literature survey; thus, their scope is limited.
    \item Some papers focus on specific types of algorithms, whereas others review general algorithms without providing detailed insights and challenges.
    \item Most of these papers do not identify major application areas where mixed data clustering is relevant.
    \item The majority of these papers ignore important practical issues such as data availability, scalability of algorithms, big data challenges, and interpretability. 
    \item Many papers do not focus on the future development of the field and does not provide guidelines to make progress.
\end{itemize}
The literature review presented in this paper attempts to avoid the drawbacks listed above and aims to contribute to the enhancement of knowledge in the field.

\section{Taxonomy for Mixed data Clustering}
\label{sec:mixed}
In recent years, there has been a surge in the popularity of mixed data clustering algorithms because many real-world datasets contain both numeric and categorical features. 
Mixed data clustering can be performed in several ways, depending on the process involved in clustering the data points. However, there exists no unified framework to structure the research being undertaken in this field. 

In this section, we  present a new taxonomy to facilitate the study of state-of-the-art mixed data clustering algorithms. This taxonomy identifies five major research themes of clustering algorithms -- \textit{partitional}, \textit{hierarchical}, \textit{model-based}, \textit{neural network-based}, and \textit{other}. The `other' category encompasses several minor groups of clustering algorithms that either do no fit into the other major research themes or have not been extensively studied. Therefore, we combine these emerging methods into a single broad research theme.  
A few clustering algorithms may belong to more than one  research themes identified by the taxonomy; however, we take great care to place them in the most appropriate thematic area of research. Table \ref{tab:groups} shows the proposed taxonomy with five different type of research themes for clustering mixed data, along with the relevant research works that is reviewed in the subsequent sections.

\begin{table}[!ht]
\centering
\caption{Taxonomy for the study of mixed data clustering algorithms.}
\label{tab:groups}
\begin{tabular}{|p{0.25cm}|p{2.65cm}|p{4cm}|}
\hline
\textbf{\#}&\textbf{Research Themes} & \textbf{Research Papers}
\\
\hline
1 & Partitional  & 
Huang \cite{Huang1997,Huang1998}, Ahmad and Dey \cite{AmirLipika2007}, Huang \textit{et al.} \cite{AutomatedWeightclus}, Modha and Spangler \cite{Modhaclustering}, Chen and He \cite{Mixeddatastream}, Ren \textit{et al.} \cite{Ren7603350}, Ji \textit{et al.} \cite{Jiimprovedkprototype}, Sangam and OM \cite{Sangam2018old}, Roy and Sharma \cite{Geneticclustering1}, Wang \textit{et al.} \cite{Wang:2015}, Wei \textit{et al.} \cite{Weie17031535}, Zhao \textit{et al.} \cite{Zhaoclsuering2007}, Chiodi \textit{et al.} \cite{Chiodiclustering}, Kaeem \textit{et al.} \cite{Mapreducemixed}, Jang \textit{et al.} \cite{GridClustering}, Barcelo-Rico and Jose-Luis \cite{Barcelo-Rico2012}, Wang \textit{et al.} \cite{Wang:2015}, Wei \textit{et al.} \cite{Weie17031535}, Cheng and Leu \cite{CHENG20095761}, Ahmad and Dey \cite{Ahmad2005}, Ji \textit{et al.} \cite{Ji2012129}, Kuri-Morales\textit{et al.} \cite{Kuri-Morales:2011:CHT:2178145.2178169},  Ji \textit{et al.} \cite{Clustercenterinitialization}, Chen \textit{et al.} \cite{Jinyin2017539}, Wangchamhan \textit{et al.} \cite{WANGCHAMHAN}, Lakhsmi et al \cite{crow}, Ahmad and Hashmi \cite{AhmadKharmonic}, Liang \textit{et al.} \cite{Liang20122251}, Yao \textit{et al.} \cite{YAO2018166}
\\ \hline
2 & Hierarchical  & 
Philips and Ottaway \cite{Philips1983}, Li and Biswas \cite{LiBiswasclustering}, Chiu \textit{et al.} \cite{Chiu:2001:RSC:502512.502549}, Hsu \textit{et al.} \cite{HsuDistanceHierarchy20074474}, , Hsu and Chen\cite{Hsu200712}, Hsu and Huang \cite{Hsu20081177}, Shih \textit{et al.} \cite{Shih2010}, Lim \textit{et al.} \cite{Lim}, Chae \textit{et al.} \cite{Chaesimilarity}
\\ \hline
3 & Model-based &
Cheeseman and Stutz \cite{Autoclass}, Everitt \cite{Everittmixedmodel}, Moustaki and Papageorgiou \cite{MoustakiRePEc:eee:csdana:v:48:y:2005:i:3:p:659-675}, Browne and McNicholas \cite{Browne20122976},  Andreopoulos \textit{et al.} \cite{BICOM},Hunt and Jorgensen \cite{Hunt3}, Lawrence and Krzanowski\cite{Lawrence1996}, McParland and Gormley \cite{McParland2016}, Sa{\^a}daoui \textit{et al.}\cite{Saadaoui}, McParland \cite{McParland}, Rajan and Bhattacharya \cite{Rajan:2016}, Tekumalla \textit{et al.} \cite{Tekumalla2017}, Marbac \textit{et al.} \cite{Marbaumodelclustering},Foss \textit{et al.} \cite{Foss2016}, Doring \textit{et al.} \cite{DoringFuzzy1336254}, Chatzis \cite{Chatzis20118684}, Pathak and Pal \cite{Pathak2016}
\\ \hline
4 & Neural network-based & 
Devaraj and Punithavalli\cite{SOM1}, Hsu \cite{Hsu:2006:GSM:2325818.2327595}, Hsu and Lin \cite{GViSOM,GsomIeee},Tai and Hsu \cite{TAI20122856}, Chen \textit{et al.} \cite{chenning},del Coso \textit{et al.} \cite{DELCOSO2015246}, Noorbehbahani \textit{et al.} \cite{Noorbehbahani2015}, Lam \textit{et al.} \cite{LamARTclustering},Hsu and Huang \cite{Hsu20081177}
\\ \hline
5 & Other &Luo \textit{et al.} \cite{Luo2006}, David and Averbuchb \cite{DAVID2012416}, Niu \textit{et al.} \cite{Niuclustering}, Ahmad and Dey \cite{Ahmadsubspace20111062}, Jia and  Cheung \cite{Subspaceclustering2}, Plant and B\"{o}hm \cite{Plant:2011:IIC:2020408.2020584}, Du \textit{et al.}  \cite{DU201746,DING2017294}, Liu et a. \cite{Liu2017}, Milenova and Campos \cite{OclusteringOracle}, Mckusick and Thompson \cite{cobweb3:a}, Reich and Fenves \cite{Reich:1991:FUA:120441.120453}, Ciaccio \textit{et al.} \cite{DiCiaccio2001}, Sowjanya and  Shashi \cite{Sowjanya}, Frey and Dueck \cite{Frey07clusteringby}, Zhang and  Gu \cite{Affinityclustering}, He \textit{et al.} \cite{Hemixed}, He \textit{et al.} \cite{Clusterensemble2005cs........9011H}, Hai and  Susumu \cite{Hai2005}, Zhao \textit{et al.} \cite{ZHAO2018264}, B\"{o}hm \textit{et al.} \cite{Bohm2010}, Behzadi \textit{et al.} \cite{parameterfree1}, Plant \cite{Plant:2012:DCA:2339530.2339589}, Li and Ye \cite{Li1597409}, Cheung and Jia \cite{CHEUNG20132228}, Sangam and Om \cite{Sangam2015}, Lin \textit{et al.} \cite{RFclusters}, Sangam and Om \cite{Sangam2018}, Yu \textit{et al.} \cite{threeway1}
\\
\hline
\end{tabular}
\end{table}

\subsection{Partitional Clustering}
\label{sec:kmean}
The most studied research theme in mixed data clustering comes from the family of partitional clustering algorithms. Most of these algorithms share characteristics with partitional algorithms developed for pure numeric data (for example K-means \cite{MacQueen1967}), or pure categorical data (for example K-modes \cite{khan2007computation}) or their variants. The general idea of these algorithms is to define
\begin{enumerate}[(i)]
    \item a cluster center that can represent categorical features and numeric features
    \item a distance measure that can combine numeric and categorical features, and
    \item a cost function, which is minimized iteratively, that can handle mixed data.
\end{enumerate}

Combining the above three ideas, most of the partitional clustering algorithms optimize the following cost function iteratively,
\begin{eqnarray}
\sum\limits_{i=1}^n\xi (d_i,C_i) 
\end{eqnarray}                    

Here, $n$ is the number of data points in the dataset,  $C_i$ is the  cluster center nearest to data point $d_i$ and $\xi$ is a distance measure between $d_i$ and $C_i$. 

An important reason for the early adoption and widespread adaptability of  partitional algorithms is that they are linear in the number of data points, scales well to large datasets and can be adapted to parallelization frameworks (for example MapReduce). Below, we review several key partitional algorithms to cluster mixed data.
 
Huang \cite{Huang1997,Huang1998} proposes the K-prototypes clustering algorithm for mixed datasets using a new cost function. New representations of cluster centers and a new definition of distance between a data point and a cluster center are proposed for mixed datasets. Cluster centers are represented by mean values for numeric features and mode values for categorical features. However, the proposed cluster center does not represent the underlying clusters well, because (i) the mode for categorical features incurs loss of information, and (ii) the Hamming distance \cite{Categoricalsimilarity} is not a good representative of the similarity between feature values for a pair of multi-valued categorical feature values. The reason is that Hamming distance gives the distance between two categorical values as only $0$ or $1$ depending upon whether two features values are same or different. Hence, this measure cannot correctly capture the distance between two differing feature values. For example, in Table \ref{tab:toy}, the Hamming distance between feature values \textit{Teaching}  and \textit{Medical} may not be the same as the distance between feature values \textit{Teaching} and \textit{Sportsman}. However, the Hamming distance will suggest otherwise and give a value of $0$ in both cases.

Ahmad and Dey \cite{AmirLipika2007} propose a new cost function and a distance measure to address these problems. They calculate the similarity between two feature values of a categorical feature from the data.
The similarity depends upon the co-occurrence of these feature values with feature values of other features. Weights of numeric features are also calculated in this method such that more significant features are given greater weights. A novel frequency-based representation of cluster center is proposed for categorical features, whereas the mean is used to for numeric features. It is shown that their proposed clustering algorithm performs better than the K-prototypes clustering algorithm.

Huang \textit{et al.} \cite{AutomatedWeightclus} extend the K-prototypes clustering algorithm to propose the W-K-prototypes clustering algorithm. In each iteration, the feature weights are updated and used in the cost function. These weights are inversely proportional to the sum of the within-cluster distances. Their results suggest an improvement in clustering results with feature weights over the clustering results achieved with the K-prototypes algorithm \cite{Huang1997,Huang1998}. Zao \textit{et al.} \cite{Zhaoclsuering2007} use the frequency of feature values for categorical features to define the cluster centers. The Hamming distance measure was used to compute the distance for categorical features, whereas mean values are used for numeric features. They show improved clustering results in comparison to the K-prototypes algorithm \cite{Huang1997,Huang1998}.

Modha and Spangler \cite{Modhaclustering} employ weighting in K-means clustering. In this method, each data point is represented in different types of feature spaces. A measure is proposed to compute the distortion between two data points in each feature space. The distortions in different feature spaces are combined to compute feature weights. The method is also employed for mixed data clustering. A mixed dataset is considered to have two feature spaces; one consisting of numeric features and the other with categorical features. Each numeric feature is linearly scaled  (by subtracting by the mean
and dividing by the standard deviation) and  1-in-q representation for each q-ary categorical feature is used. The squared Euclidean distance is used for numeric features, whereas the cosine distance is used for categorical features. No comparative study with other clustering algorithms is presented in the paper.

Chen and He \cite{Mixeddatastream} use the distance measure suggested by Ahmad and Dey \cite{AmirLipika2007} to propose a data clustering algorithm for  data streams with mixed numeric and categorical features.  The concept of micro-clusters is used in the algorithm. Micro-clusters are used to compress the data efficiently in data streams. In the first stage, initial cluster centers are calculated to cluster the data. The method uses two parameters: decay factor and dense threshold. Decay factor defines the significance of historical data to the current cluster whereas the dense threshold is used to distinguish between dense and sparse micro-clusters. The parameter optimization is a potential problem with the method. 

Ran \textit{et al.} \cite{Ren7603350} use the cluster centers proposed by Ahmad and Dey \cite{AmirLipika2007} to develop another mixed data clustering algorithm. Euclidean distance for numeric features and Hamming distance for categorical features are used to compute the similarity between the cluster center and a data point, with a Gaussian kernel function applied to the total distance. Ji \textit{et al.} \cite{Jiimprovedkprototype} combine the definition of cluster center \cite{AmirLipika2007} with the significance of feature  \cite{AutomatedWeightclus} to propose a new cost function.  The significance of a feature is initially selected randomly, followed by an update to its value with each iteration. The random selection of  the significance of a feature can worsen the problem of random initialization of the cluster center  \cite{kmeanclusterinitialization1,kModeclusterinitilization1}  because it would lead to different results in different runs.

Sangam and Om \cite{Sangam2018old} propose a new distance measure for the K-prototypes clustering algorithms. The weightage Hamming distance is proposed for categorical features, this is based on the frequency of feature values in different clusters. The Minkowski distance measure is used to compute the distance for numeric features. The proposed method outperforms the original K-prototypes clustering algorithm.

Roy and Sharma \cite{Geneticclustering1} extend the fast genetic K-means clustering technique (FGKA) \cite{Lu} for mixed data. The algorithm minimizes the total within-cluster variation.  They use the distance measure proposed by Ahmad and Dey \cite{AmirLipika2007} in their algorithm. They claim that the algorithm performs better than the FGKA algorithm \cite{Lu}; however, they do not explain the modification made in FGKA (which can handle only numeric data) to allow  mixed data.

Chiodi \textit{et al.} \cite{Chiodiclustering} propose an iterative partitional clustering algorithm for mixed data, which is motivated by the K-means clustering algorithm. They propose a cost function which computes the mean diversity of the data points in a cluster with respect to all of the features. The Euclidean distance measure is used for a numeric feature and the Hamming distance is used for categorical features. Mean values are used for numeric features and the frequency distribution is used for categorical values in clusters. The algorithm is applied to the andrological dataset.  Kacem \textit{et al.} \cite{Mapreducemixed} propose parallelization of the K-prototypes clustering method \cite{Huang1998} to handle large mixed datasets, this algorithm uses the MapReduce framework \cite{Dean:2008:MSD:1327452.1327492} for parallelization. Jang \textit{et al.} \cite{GridClustering} use a grid-based indexing technique to develop grid-based K-prototypes algorithm that speeds up K-prototypes algorithm. The experiments carried out using a spatial dataset consisting of numeric and categorical features show that the proposed method takes less time than the original K-prototypes algorithm.
Table \ref{tab:kmean} summarizes different K-means-type algorithms for mixed data clustering.

\begin{table*}[!ht]
\centering
\caption{K-means-type clustering  algorithms for mixed datasets.}
\label{tab:kmean}
\begin{tabular}{|p{3cm}|p{6cm}|p{5.5cm}|}
\hline
\textbf{Algorithm} & \textbf{Center Definition}  & \textbf{Distance  Measure}
\\
\hline
Huang \cite{Huang1997,Huang1998}&Mean values for numeric features, mode values for categorical data & Euclidean distance for numeric features, Hamming distance for categorical features
\\Ahmad and Dey \cite{AmirLipika2007}&Mean values for numeric features, proportional frequency-based center for categorical features & Weights for numeric features are calculated,  Euclidean distance for numeric features and   co-occurrence-based distance measure for categorical features 
\\ Huang \textit{et al.}  \cite{AutomatedWeightclus}&Mean values for numeric features, mode values for categorical features &  Weights of features based on the importance of the features in clustering are calculated in each run with distance measure used by Huang \cite{Huang1997,Huang1998} 
\\Zhao \textit{et al.} \cite{Zhaoclsuering2007} &Mean values for numeric features, proportional frequency-based center for categorical features & Euclidean distance for numeric features, Hamming distance for categorical features 
\\Modha and Spangler \cite{Modhaclustering} & First, 1-in-q representation for each q-ary categorical feature, Mean values for all features &Weights of features are calculated, squared Euclidean distance is used for numeric features whereas cosine distance is used for categorical features 
\\Ji \textit{et al.} \cite{Jiimprovedkprototype} & Center as proposed by Ahmad and Dey \cite{AmirLipika2007} & Weights are calculated by the method suggested by Huang \textit{et al.} \cite{AutomatedWeightclus}, squared Euclidean distance is used for numeric features, Hamming distance is used for categorical features 
\\ Ran \textit{et al.} \cite{Ren7603350} &Center as proposed by Ahmad and Dey \cite{AmirLipika2007}  & Gauss kernel function 
\\
\hline
\end{tabular}
\end{table*}

The other partitional approach to mixed data clustering is to first convert a mixed dataset into a numeric dataset and then apply traditional K-means clustering to it. Barcelo-Rico and Jose-Luis \cite{Barcelo-Rico2012} develop a method that uses polar or spherical coordinates to codify categorical features into numeric features and then uses K-means clustering on the new numeric datasets. Their method outperforms K-modes clustering algorithms and K-prototypes clustering method. Wang \textit{et al.} \cite{Wang:2015}  propose the context-based coupled representation for mixed datasets. The interdependence of numeric features and the interdependence of categorical features are computed separately and then, the interdependence across the numeric and categorical features is computed. These relationships form the numeric  representation for mixed-type data points. The K-means clustering algorithm is used to cluster these new data points. Their experimental results suggest that the method outperform other mixed-data clustering algorithms. Wei \textit{et al.} \cite{Weie17031535} propose a  mutual information-based transformation method for unsupervised features that can convert categorical features into numeric features, which are then clustered by using K-means clustering algorithm. Table \ref{tab:mixed2numeric} summarizes the clustering methods that first convert the mixed data to numeric data and then apply the K-means clustering  on the new numeric data. 

Constraint-based clustering \cite{Wagstaff:2001:CKC:645530.655669}  groups similar data points into several clusters under certain user constraints: for example. that two given data points should belong to the same cluster. 
Cheng and Leu \cite{CHENG20095761} propose a constrained K-prototypes clustering algorithm that simultaneously handles user constraints and mixed data. The algorithm extends the K-prototypes clustering algorithm \cite{Huang1998} by adding a constrained function to the cost function of the K-prototypes.

\begin{table*}[!ht]
\centering
\caption{Clustering algorithm when categorical features are converted to numeric features.}
\label{tab:mixed2numeric}
\begin{tabular}{|p{4.5cm}|p{8.5cm}|}
\hline
\textbf{Algorithm} & \textbf{Method to convert the categorical features to numeric features} 
\\
\hline
Barcelo-Rico and Jose-Luis \cite{Barcelo-Rico2012} & Coding is based on polar or spherical coordinates
\\ Wang \textit{et al.} \cite{Wang:2015} & Context-based coupled relationship for mixed data
\\  Wei \textit{et al.} \cite{Weie17031535} & Mutual information (MI)-based unsupervised feature transformation
\\
\hline
\end{tabular}
\end{table*}

Fuzzy clustering represent those approaches in which a data point can belong to more than one cluster with different degrees (or probabilities) of membership \cite{fuzzyYANG19931}. Various fuzzy clustering algorithms have been proposed for mixed data based on partitional clustering. Ahmad and Dey \cite{Ahmad2005} use a dynamic probabilistic distance measure to determine the weights of numeric features and 
distances between each pair of categorical values of a categorical feature. The distance measure is combined with
the cluster center definition suggested by El-Sonbaty and Ismail
\cite{El-Sonbaty} to develop a fuzzy C-means (FCM) clustering algorithm \cite{fuzzycmenas1,fuzycmenas2} for mixed
data. Ji \textit{et al.} \cite{Ji2012129}  propose a fuzzy  clustering method for mixed data by combining the similarity measure proposed by Ahmad and Dey \cite{AmirLipika2007} with the cluster center definition suggested by El-Sonbaty and Ismail
\cite{El-Sonbaty}. Kuri-Morales\textit{et al.} \cite{Kuri-Morales:2011:CHT:2178145.2178169}  propose a strategy for the assignment of a numeric value to a categorical value. First, a mixed dataset is converted into a pure numeric dataset and then fuzzy C-means clustering algorithm is used.

Partitional clustering algorithms for numeric and categorical data (for example K-means and K-modes) suffer from several drawbacks, such as cluster center initialization \cite{kmeanclusterinitialization1,kModeclusterinitilization1} and the prior knowledge of the number of clusters \cite{MacQueen1967}. Because of their conceptual similarity, these issues also exist in their counterparts for mixed datasets.
In the next subsections, we review relevant literature that covers these  issues.

\subsubsection{Cluster Center Initialization}
Cluster center initialization is a well-known problem with partitional clustering algorithms \cite{kmeanclusterinitialization1,kModeclusterinitilization1,khan2003computing}. In these algorithms, initial cluster centers are usually selected randomly this may lead to different clustering outcomes on different runs of the algorithm. Therefore, data mining researchers may find it 
difficult to rely on such clustering outcomes.

Ji \textit{et al.} \cite{Clustercenterinitialization} propose an algorithm to create initial cluster centers for K-means-type algorithms for mixed datasets. They introduce the idea of the centrality of data points, which uses the concept of neighbor-set. The centrality and distances are used to compute the initial cluster centers. However, their algorithm has quadratic complexity, in contrast to the linear time complexity of K-means-type clustering algorithms.

Using density peaks \cite{Rodriguez1492},
Chen \textit{et al.} \cite{Jinyin2017539} propose an algorithm to determine the initial cluster centers for mixed datasets. Higher-density points are used
to identify cluster centers. This algorithm has quadratic complexity, hence, it is not useful for K-means-type algorithms.  Wangchamhan \textit{et al.} \cite{WANGCHAMHAN} combine a search algorithm, League Championship Algorithm \cite{LeagueChampionshipAlgorithm}, with the K-means clustering algorithm to identify the initial cluster centers. They apply Gower's distance measure \cite{Gowersimilarity} to find the distance between a data point and a cluster center. Parameter selection is a problem with this approach. 
Lakhsmi et al \cite{crow} uses the crow optimization method to compute the initial cluster centers for the K-prototypes clustering algorithm. This algorithm outperforms the K-prototypes clustering algorithm with random initial cluster centers. The selection of parameters in crow optimization is an important step; the same clustering results may not be produced by using different parameters. 

Ahmad and Hashmi \cite{AhmadKharmonic} combine the distance measure and the definition of centers for mixed data  proposed by Ahmad and Dey \cite{AmirLipika2007} with the cost function of K-harmonic clustering \cite{ZhangKHarmonic} to extend K-harmonic clustering to mixed data. Their results indicate that their method is robust to the selection of initial cluster centers as compared to other K-means clustering type algorithms for mixed datasets. Zheng et a. \cite{Zheng5586136} combine an evolutionary algorithm (EA) with the K-prototypes clustering algorithm \cite{Huang1998}. The global
searching ability of EA makes the proposed algorithm less sensitive to cluster initialization.

\subsubsection{Number of Clusters}
Most of the partitional clustering algorithms for numeric and categorical data work under the assumption that the number of clusters is known in advance. This number may be either computed by other algorithms, derived from the domain, or user-defined. However, many of these methods may not guarantee that the chosen number of clusters corresponds to the natural number of clusters in the data. The same problem exists for partitional algorithms for mixed data.

Liang \textit{et al.} \cite{Liang20122251} propose a cluster validity index to discover the number of clusters for mixed data clustering. This index has two components: one for numeric features and the other for categorical features. For categorical features, the cluster validity index uses the category utility function developed by Gluck and Corter \cite{Gluck}. For numeric features, a corresponding category utility function proposed by Mirkin \cite{Mirkin2001} is used. Each component is given a weight depending upon the number of categorical and numeric features and the total number of features. The cluster validity index is computed for different number of clusters. The number of clusters that maximizes the cluster validity index is chosen as the optimal number of clusters. In this method, the process starts with a large number of clusters and in each round the worst cluster is combined with other clusters. Renyi entropy \cite{Renyi} for numeric features and complement entropy \cite{Langcomplement} for categorical features are used to determine the worst cluster. The method is used with the K-prototypes method \cite{Huang1998}. The algorithm is successful in finding the number of clusters in various datasets. These datasets have predefined classes and the number of the classes is taken as the number of clusters in the datasets.   Yao \textit{et al.} \cite{YAO2018166} extend  the algorithm \cite{Liang20122251} by adding a method to find the initial clusters to avoid the cluster initialization problem. However, the method to find initial clusters is based on density estimation which makes the method quadratic. The comparative study suggests that the original method \cite{Liang20122251} may produce different number of clusters in different runs whereas the proposed method produces the same number of clusters. The experiment shows that the method is successful in predicting the correct number of clusters in datasets. 

Rahman and Islam \cite{Rahman2014345} combine genetic algorithm optimization  \cite{GABook} and the K-means clustering algorithm to produce a clustering algorithm for mixed data that computes the number of clusters automatically. They use the distance measure proposed by Rahman and Islam \cite{Rahman2012} to compute the distance between a pair of categorical values. The algorithm shows good results; however, its complexity is quadratic.

\subsection{Hierarchical Clustering}
Hierarchical clustering methods create a hierarchy of clusters organized in a top to down (or bottom to up) order.
To create clusters, the hierarchical algorithms need both of the following:
\begin{enumerate}[(i)]
    \item Similarity matrix  - This is constructed by finding the similarity between each pair of mixed data points. The choice of similarity metric (to construct a similarity matrix) influences the shape of the clusters,
    \item Linkage criterion -- This determines the distance between sets of observations as a function of the pairwise distances between observations.
\end{enumerate}
Most hierarchical clustering algorithms have a large time complexity of $O(n^3)$ and requires $O(n^2)$ memory, where $n$ is the number of data points. Below, we review several hierarchical clustering algorithms that have been developed to handle mixed data.

Philip and Ottaway \cite{Philips1983} use Gower's similarity measure \cite{Gowersimilarity} to compute the similarity matrix for mixed datasets. Gower's similarity measure computes the similarity by dividing features into two subsets one for categorical features and the other for numeric features. Hamming distance is applied to compute the similarity between two data points for a categorical feature. A weighted average of similarities for all categorical features is the similarity between two data points in a categorical feature space. For numeric features, the similarity is computed such that the similarity between the same feature values is $1$, whereas if the difference between the values is the maximum possible difference (the difference between maximum and minimum values of the feature) the similarity is $0$.   
The sum of the similarity values for all numeric features is the similarity for two data points in a numeric feature space. The similarity in the categorical feature space and the numeric feature space are added to compute the similarity between two data points. Hierarchical agglomerative clustering is then used to create clusters. 
Chiu \textit{et al.} \cite{Chiu:2001:RSC:502512.502549} develop a similarity measure to compute the similarity between two clusters for mixed data. This similarity measure is related to the decrease in the log-likelihood function when two clusters are
merged. The authors combine the  BIRCH clustering algorithm \cite{BirchZhang1997}, which uses hierarchical clustering algorithm, with their proposed similarity measure to develop a clustering algorithm that can handle mixed datasets. 
Li and Biswas \cite{LiBiswasclustering} propose similarity-based agglomerative clustering (SBAC) algorithm for mixed data clustering. SBAC uses the Goodall similarity measure \cite{Goodallsimilarity} and applies a hierarchical agglomerative approach to build cluster hierarchies. 

Hsu \textit{et al.} \cite{HsuDistanceHierarchy20074474} propose a distance measure based on  a concept hierarchy consisting of concept nodes and links \cite{Conceptclustering1,Conceptclustering2}. The more general concepts are represented by higher-level nodes, whereas more specific concepts are represented by lower-level nodes. The categorical values are represented by a  tree structure such that each leaf is represented by a categorical value. Each feature of a data point is associated with a distance hierarchy. The distances between two data points is calculated by using their associated distance hierarchies. 
An agglomerative hierarchical clustering algorithm \cite{ClusteringbookJain} is applied to a distance matrix to obtain the clusters. Domain knowledge is required to make distance hierarchies for categorical features, and is non-trivial in many cases.
Hsu and Chen \cite{Hsu200712} propose a new similarity measure to cluster mixed data. The algorithm uses variance
for computing the similarity of numeric values. For similarity between
categorical values, they \cite{Hsu200712} utilizes 
entropy with distance hierarchies
\cite{HsuDistanceHierarchy20074474}. The similarities are then aggregated to compute the similarity matrix for a mixed dataset. Incremental clustering is used on the similarity matrix to obtain the clusters. In an extended work, Hsu and Huang \cite{Hsu20081177} apply an adaptive
resonance theory network (ART) to cluster data points by using the distance hierarchies as the input of the network. A better interpretation of clusters is possible with the proposed algorithm as compared to the K-prototypes algorithm. Shih \textit{et al.} \cite{Shih2010} convert categorical features of a mixed dataset into numeric features by using frequencies of co-occurrence of categorical feature values. The dataset with all numeric  features is then clustered by using a hierarchical agglomerative clustering algorithm \cite{ClusteringbookJain}.

Lim \textit{et al.} \cite{Lim} partition the data into two parts:  categorical data and numeric data. The two types of data are clustered separately. The clustering results are combined by using a weighted scheme to obtain a similarity matrix. The agglomerative hierarchical clustering method is applied on the similarity matrix to obtain the final clusters.
Gower's similarity measure assigns equal weights to both types of features in computing the similarity between two data points. The similarity matrices may be dominated by one feature type.  
Chae \textit{et al.} \cite{Chaesimilarity}  assign weights to the different feature types to overcome this problem. Improved clustering results are shown with these weighted similarity matrices.

Table \ref{tab:hierarchical} summarizes the different hierarchical clustering methods for mixed data that were discussed in this section.

\begin{table*}
\centering
\caption{Hierarchical clustering algorithms for mixed datasets.}
\label{tab:hierarchical}
\begin{tabular}{|p{3cm}|p{7cm}|p{4cm}|}
\hline
\textbf{Algorithm} & \textbf{Similarity measure for a similarity matrix} & \textbf{Clustering algorithm}  
\\
\hline
Philip and Ottaway \cite{Philips1983} & Gower's similarity Matrix \cite{Gowersimilarity} &Agglomerative hierarchical clustering method
\\  Chiu \textit{et al.} \cite{Chiu:2001:RSC:502512.502549} & Probabilistic model by using a log-likelihood function & BIRCH algorithm  \cite{BirchZhang1997}
\\ Li and Biswas \cite{LiBiswasclustering} & Goodall similarity measure \cite{Goodallsimilarity} & Agglomerative hierarchical clustering with group- average method
\\ Hsu \textit{et al.} \cite{HsuDistanceHierarchy20074474} & Distance hierarchy by using concept hierarchy \cite{Conceptclustering1,Conceptclustering2} & Agglomerative hierarchical clustering
\\ Hsu and Chen \cite{Hsu200712} & Variance for numeric features and entropy with distance hierarchies
\cite{HsuDistanceHierarchy20074474} for categorical features  & Incremental clustering 
\\ Hsu and Huang \cite{Hsu20081177} & Similarity measure proposed by Hsu and Chen \cite{Hsu200712} & Adaptive
resonance theory network \cite{ART}
\\Shih \textit{et al.} \cite{Shih2010} & Convert categorical features  into numeric features & Hierarchical agglomerative clustering algorithm  \cite{ClusteringbookJain}
\\Lim \textit{et al.} \cite{Lim} & Two similarity matrices: one for categorical data and one for numeric data & Agglomerative hierarchical clustering method 
\\Chae \textit{et al.} \cite{Chaesimilarity} & Modified Gower's similarity matrix &Agglomerative hierarchical clustering method
\\
\hline
\end{tabular}
\end{table*}

\subsection{Model-based clustering}
Model-based clustering methods assume that a data point matches a model, which in many cases, is a statistical distribution \cite{Modelbasedclustering}. The models are generally user-defined, so they are prone to yielding undesirable clustering outcomes if inappropriate models (or their parameters) are chosen.
Model-based clustering algorithms are generally slower than partitional algorithms \cite{Modelbasedclustering}. Next, we review several model-based clustering algorithms for mixed data.

AUTOCLASS \cite{Autoclass} performs clustering by integrating finite mixture distribution and Bayesian methods with  prior distribution of each feature. AUTOCLASS can cluster data containing both categorical and numeric features.
Everitt \cite{Everittmixedmodel} proposes a clustering algorithm by using model-based clustering for datasets consisting of both numeric
features and binary or ordinal features.  The normal model is extended to handle  mixed datasets by  using thresholds for the categorical features. Because of high computational cost, the method is only useful for datasets containing very few categorical features. To overcome this problem, Lawrence and Krzanowski \cite{Lawrence1996} extend the homogeneous Conditional Gaussian model to the finite mixture case, to compute maximum likelihood estimates for the parameters in a sample population. They suggest that their method works for  an arbitrary number of features.

Moustaki and Papageorgiou \cite{MoustakiRePEc:eee:csdana:v:48:y:2005:i:3:p:659-675} use
a latent class mixture model for mixed data clustering. Categorical features are converted to binary features by a 1-in-q representation. A multinomial distribution is used for categorical features and a normal distribution is used for a numeric features. Features are considered independent in each cluster. The algorithm has been applied to an archaeological dataset. Browne and McNicholas \cite{Browne20122976} propose a mixture of latent features model for clustering,  the expectation-maximization (EM) framework \cite{Dempster77maximumlikelihood} is used for model fitting. Andreopoulos \textit{et al.} \cite{BICOM} present a clustering algorithm, Bi-level clustering of mixed categorical and numeric data types (BILCOM) for mixed datasets. The algorithm uses categorical data clustering to guide the clustering of numeric data.  Hunt and  Jorgensen \cite{Hunt3,Hunt12003429,Hunt2WIDM:WIDM33} propose a mixture model clustering approach for mixed data. In this approach, a finite mixture of multivariate distributions is fitted to data and then the membership of each data point is calculated by computing the conditional probabilities of cluster membership. A local independence assumption  can be used to reduce the model parameters. They  further show that the method can also be applied for clustering mixed datasets with missing values \cite{Hunt12003429}.

The ClustMD method \cite{McParland2016} uses a latent variable model to cluster mixed datasets.  It is suggested that a latent variable with a mixture of Gaussian distributions produces the observed mixed data. An EM algorithm is applied to estimate the parameters for ClustMD. A Monte Carlo EM algorithm \cite{MonteEM} is used for datasets having categorical features. This method can model both numeric and categorical features; however, it becomes computationally expensive as the number of features increases. To overcome this problem, McParland \textit{et al.} \cite{McParlandmodelclustering} propose a clustering algorithm for high-dimensional mixed data by using a Bayesian finite mixture model. In this algorithm, the estimation is done by using the Gibbs sampling algorithm. To select the optimal model, they also propose an approximate Bayesian Information Criterion-Markov chain Monte Carlo criterion. They show that the method works well on a mixed medical dataset consisting of high-dimensional numeric phenotypic features and categorical genotypic features. Saadaoui \textit{et al.} \cite{Saadaoui} propose a projection of
the categorical features on the subspaces spanned by numeric features; an optimal Gaussian mixture model is obtained from the resulting principal component analysis regressed subspaces.

 Rajan and Bhattacharya \cite{Rajan:2016} present a clustering algorithm based on  Gaussian mixture copulas\footnote{Copulas are defined as ``functions that join or
couple multivariate distribution functions to their one-dimensional
marginal distribution functions'' and as ``distribution functions whose
one-dimensional margins are uniform.'' \cite{Nelsen:2006:IC:1204326}.}
that can model dependencies between features and can be applied to datasets having numeric and categorical features. Their method outperforms other clustering algorithms on a variety of datasets.  Tekumalla \textit{et al.} \cite{Tekumalla2017} use the concept of vines copulas\footnote{Vine copulas provide a flexible way of pair-wise dependency modeling using hierarchical collections of bivariate copulas, each of which can belong to any copula family thereby capturing a wide variety of dependencies \cite{Tekumalla2017}.} 
for mixed data clustering, they propose an inferencing algorithm to fit those vines on the mixed data. A dependency-seeking multi-view clustering that uses a Dirichlet process mixture of vines is developed \cite{Tekumalla2017}. Marbac \textit{et al.} \cite{Marbaumodelclustering} present a mixture model of Gaussian copulas for mixed data clustering. In this model, a component of the Gaussian copula mixture creates a correlation coefficient for a pair of features. They select the model by using two information criteria: the Bayesian information criterion  \cite{schwarz1978} and  integrated completed likelihood  criterion \cite{ICLcriterion}. The Bayesian inference is performed by using a Metropolis-within-Gibbs sampler. Foss et al \cite{Foss2016} develop a semi-parametric method, KAy-means for MIxed LArge data (KAMILA), for clustering mixed data. KAMILA balances the effect of the numeric and categorical features on clustering.  KAMILA integrates two different kinds of clustering
algorithms; the K-means algorithm  and Gaussian-multinomial mixture models \cite{Hunt2WIDM:WIDM33}. Like the K-means clustering algorithm, no strong parametric assumptions are made for numeric features in the KAMILA algorithm. KAMILA uses the properties of Gaussian-multinomial mixture models to balance the effects of numeric and categorical features without specifying weights.

Doring \textit{et al.} \cite{DoringFuzzy1336254} propose a fuzzy clustering algorithm for mixed data by using a mixture model. The mixture model is used to determine the similarity
measure for mixed datasets.  It also helps in  finding the cluster prototypes. The inverse of the probability that a data point occurs in a cluster is used to define the distance between the cluster center and the data point. Chatzis \cite{Chatzis20118684} proposes a FCM-type clustering algorithm for mixed data that employs a probabilistic dissimilarity function in a FCM-type fuzzy clustering cost function  proposed by Honda and Ichihashi \cite{Honda1492403}. Pathak and Pal \cite{Pathak2016} combine  fuzzy, probabilistic and collaborative clustering in a  framework for mixed data clustering. Fuzzy clustering is used to cluster numeric data portion of the mixed data, whereas mixture models \cite{Booksmachine,Chatzis20118684} are used to cluster categorical data portion. Collaborative clustering \cite{Pedrycz2002} is used to find the common cluster sub-structures in the categorical and numeric data.

 Table \ref{tab:model} summarizes the various model-based clustering algorithms for mixed data that are discussed in this section.

\begin{table*}
\centering
\caption{Model-based clustering algorithms for mixed datasets.}
\label{tab:model}
\begin{tabular}{|p{5cm}|p{10cm}|}
\hline
\textbf{Algorithm} &\textbf{Model}
\\
\hline
Cheeseman and Stutz \cite{Autoclass} & Bayesian methods  
\\  Everitt \cite{Everittmixedmodel} & Model-based clustering with the use of thresholds for the categorical features.
\\Lawrence and Krzanowski \cite{Lawrence1996} & Extension of homogeneous conditional Gaussian model to the finite mixture situation.
\\Moustaki and Papageorgiou \cite{MoustakiRePEc:eee:csdana:v:48:y:2005:i:3:p:659-675} & Latent class mixture model.
\\Browne and McNicholas \cite{Browne20122976} & A mixture of latent variables model with the expectation-maximization framework. \cite{Dempster77maximumlikelihood}.
\\ Andreopoulos \textit{et al.} \cite{BICOM} &  Pseudo-Bayesian process with categorical data clustering to guide the clustering of numeric data.
\\Hunt and  Jorgensen \cite{Hunt3,Hunt12003429,Hunt2WIDM:WIDM33}& A finite mixture of multivariate distributions is fitted to data.
\\McParland and Gormley \cite{McParland2016} & A latent variable model.
\\McParland \textit{et al.} \cite{McParland} & Bayesian finite mixture model.
\\Saadaoui \textit{et al.} \cite{Saadaoui} & A projection of
the categorical features on the subspaces spanned by numeric features and then the application of Gaussian Mixture Model.
\\Rajan and Bhattacharya \cite{Rajan:2016}&   Gaussian mixture copula. 
\\Tekumalla1 \textit{et al.} \cite{Tekumalla2017} &  Vine copulas and Dirichlet process mixture of vines. 
\\ Marbac \cite{Marbaumodelclustering} &A mixture model of Gaussian copulas.
\\ Foss \textit{et al.} \cite{Foss2016} &K-means algorithm  and Gaussian-multinomial mixture models
\\
\hline
\end{tabular}
\end{table*}

\subsection{Neural network-based clustering}
Most of the research on clustering mixed data using neural networks is focused on using self organizing maps (SOM) \cite{Kohonen1982} and adaptive  resonance  theory (ART)  \cite{LamARTclustering} approaches. A SOM \cite{Kohonen1982,Kohonen:2001:SM:558021} is a neural network that is used to non-linearly project a dataset onto a lower-dimensional feature space so that cluster analysis can be performed in the new feature space. 
ART is based on the theory of how the brain learns to categorize autonomously and predict in a dynamic world \cite{grossberg2013adaptive}. The key aspect of ART's predictive power is its ability to carry out fast, incremental, and stable unsupervised and supervised learning in response to a changing world \cite{grossberg2013adaptive}. Both the traditional SOM-based and ART-based clustering methods can handle numeric features, however they cannot be used directly for categorical features. Categorical features are first transformed into binary features, which are then treated as numeric features \cite{SOM1,LamARTclustering}. 

Hsu \cite{Hsu:2006:GSM:2325818.2327595} develops a generalized SMO model to compute the similarity of categorical values by using a distance hierarchy that is based on a concept hierarchy. It consists of nodes and weighted links: more general concepts are represented by higher-level nodes whereas more specific concepts are represented by lower-level nodes. Distance hierarchies are also used to compute the similarities between two data points in the complete (numeric and categorical) feature space.
Visualization-induced SMO \cite{Yin:2002:VNM:2325783.2326851} preserves the structure of data in the new low-dimensional space better than SMO. Hsu and Lin \cite{GViSOM} combine generalized SMO with visualization-induced SMO to develop a method generalized visualization-induced SOM to cluster mixed datasets. The experiments suggest that the method gives excellent cluster analysis results. Hsu and Lin \cite{GsomIeee} modify the distance measure presented in Generalized SMO and use the Visualization-Induced SMO to develop a new method for mixed data clustering. Traditional SMO has the weakness that it has predefined fixed-size map; to improve its flexibility,  growing SMO is proposed \cite{Alahakoon:2000:DSM:2325773.2326577}. Growing SMO starts with a small map that grows with training data. Tai and Hsu \cite{TAI20122856} integrate generalized SMO with growing SMO to develop a clustering algorithm for mixed datasets. Chen and Marques \cite{chenning} propose a clustering algorithm based on SMO,  using  the Hamming distance for categorical features and the Euclidean distance for numeric features. This method has the problem that it gives more weight to categorical features, to overcome this problem Coso \textit{et al.} \cite{DELCOSO2015246} modify the distance measure such that each type of feature has equal weight. The method show better results than the method presented by Chen and Marques. 
Noorbehbahani \textit{et al.} \cite{Noorbehbahani2015} propose an incremental mixed-data clustering algorithm which uses a self-organizing incremental neural network algorithm\cite{Furao200690}. They also propose a new distance measure in which the distance between two categorical values depend on the frequencies of those features. The co-occurrence of feature values \cite{AmirLipika2007}, which may affect the accuracy of the distance measure, is not considered.

Lam \textit{et al.} \cite{LamARTclustering} use an unsupervised feature learning approach to obtain a sparse representation of  mixed datasets. A fuzzy adaptive resonance theory (ART) approach \cite{Carpenter1991759} is used to create new features. First, fuzzy ART approach is used to create prototypes of the dataset, which are then employed as mixed features encoder to map individual data points to the new feature space. They use K-means clustering algorithm to cluster data points in the new feature space.  Hsu and Huang \cite{Hsu20081177} use ART to create a similarity matrix that can be used to cluster data points by using hierarchical clustering.

\subsection{Other} 
In the previous sections, we summarized major contributions on the four prominent research themes adopted by researchers for clustering mixed data. However, several new sub-themes and research directions have emerged in recent years. As many of these new research directions have not been explored enough, we combine them under one umbrella theme named `\textit{Other}'. Many of these new types of clustering algorithms may not fit within the realms of the more established research themes as discussed in previous sections. 

Spectral clustering techniques \cite{Ng:2001:SCA:2980539.2980649} perform dimensionality reduction by using eigenvalues of the similarity matrix of the data. Thereafter, the clustering is performed in fewer dimensions. First a similarity matrix is computed, and then a spectral clustering algorithm \cite{Ng:2001:SCA:2980539.2980649} is applied to this similarity matrix to obtain clusters.
Luo \textit{et al.} \cite{Luo2006} propose a similarity measure by using a clustering ensemble technique. In this measure, the similarity of two data points is computed separately for numeric and categorical features. The two similarities are added to obtain the similarity between two data points. Spectral clustering is used on the similarity matrix to obtain the clusters.   
David and Averbuchb \cite{DAVID2012416} propose a clustering algorithm, SpectralCAT, which uses categorical spectral clustering to cluster mixed datasets.
The algorithm automatically transforms the numeric features to categorical values. 
This is performed by finding the optimal transformation according to the Calinski and Harabasz index \cite{Calinski}. A spectral clustering method is then applied to the transformed data \cite{DAVID2012416}. Niu \textit{et al.} \cite{Niuclustering} present a clustering algorithm for mixed data, in which the similarity matrices for numeric  and categorical features are computed separately. Coupling relationships of features are used to compute similarity matrices. Both matrices are combined by  weighted summation to compute the similarity matrix for the mixed data. This algorithm is applied to find the clusters for a web-based learning system data, The results suggest that the method outperforms the K-prototypes clustering algorithm and the SpectralCAT algorithm \cite{DAVID2012416}.

Subspace clustering \cite{Subspaceclusteringdefinition} seeks to discover clusters in different subspaces within
a dataset.
Ahmad and Dey \cite{Ahmadsubspace20111062} use a distance measure \cite{AmirLipika2007} for the mixed data with a cost function for subspace clustering \cite{Jing:2007} to develop a K-means-type clustering algorithm, which can produce subspace clustering of mixed data.
Jia and  Cheung \cite{Subspaceclustering2} present a feature-weighted clustering model that uses data point-cluster similarity for soft subspace clustering of mixed datasets. They propose a unified
weighting scheme for the numeric and categorical features, which determines the feature-to-cluster contribution. The method finds the most appropriate number of clusters automatically. Plant and B\"{o}hm \cite{Plant:2011:IIC:2020408.2020584} develop a clustering technique, interpretable clustering of numeric and categorical objects (INCONCO), which produces interpretable clustering results for mixed data. The algorithm uses the concept of data compression by using the  minimum description length (MDL) principle \cite{MDLRISSANEN1978465}. INCONCO identifies the relevant feature dependencies using linear models and provides subspace clustering  for mixed datasets. INCONCO does not support all types of feature dependencies. The algorithm  demands that all values of categorical features involved in a dependency with some numeric features must have a unique numeric data distribution.

Density-based clustering methods assume that clusters are defined by dense regions in the data space, separated by less dense regions \cite{DBSCAN}.
Du \textit{et al.}  \cite{DU201746,DING2017294} propose a new distance measure for mixed data clustering, in which they assign a weight to each categorical feature. They combine this distance measure with a density peaks clustering algorithm \cite{Rodriguez1492} to cluster mixed datasets. However, the selection of different parameters makes it difficult to use in practice.
Liu \textit{et al.} \cite{Liu2017} propose a density-based clustering algorithm for mixed datasets. The authors extend 
the DBSCAN algorithm \cite{DBSCAN} to mixed datasets. Entropy is used to compute the distance measure for mixed datasets. 
Milenova and Campos \cite{OclusteringOracle} use orthogonal projections to cluster mixed datasets. These orthogonal projections are used to find high-density regions in the input data space. Du \textit{et al.} \cite{Du8367173} propose a density-based clustering method for mixed datasets. Datasets can be divided into three categories depending upon the ratio of the number of categorical features and the number of numeric features. Different mathematical models are suggested for these categories. First, numeric features are used to create clusters  categorical features are used to create clusters, and finally, these clusters are combined to obtain the final clusters. 

Conceptual clustering \cite{Fisher1987} generates a concept description for each generated cluster. Generally, conceptual clustering methods generate  hierarchical category structures. COBWEB \cite{Fisher1987} uses a category utility (CU) measure \cite{Gluck} to define the  relation between groups or clusters. As the CU measure can only handle categorical features, the CU measure is extended to handle numeric features for mixed data clustering. COBWEB\/3 \cite{cobweb3:a} integrates the original COBWEB algorithm with the method presented in CLASSIT \cite{GENNARI198911}  to deal with numeric features in the CU measure. With this method, it is assumed that numeric feature values are normally distributed. To overcome the problem of normal distribution assumption, a new method ECOBWEB \cite{Reich:1991:FUA:120441.120453}, which uses the probability distribution of the average value for a feature, is presented.

Ciaccio \textit{et al.} \cite{DiCiaccio2001} extend the well-separated partition definition \cite{Jardineclustering} to propose a non-hierarchical clustering algorithm for mixed data, which can analyze large amount of data in the presence of missing values.
Sowjanya and  Shashi \cite{Sowjanya} propose an incremental clustering approach for mixed data. Initially, some data points are clustered and other data points are assigned to clusters depending upon their distances from the cluster centers, which are updated as new data points join the clusters. A cluster center is defined, for a categorical feature, by using the mode of the categorical values of data points present in the cluster. For a numeric feature, the mean of the values of the data points present in a cluster  is used to represent the center of the cluster. However, it is not clear in the paper which distance measure is used to cluster data points.

Frey and Dueck \cite{Frey07clusteringby} propose an affinity propagation clustering (APC) algorithm that uses message passing. Zhang and  Gu \cite{Affinityclustering} extend this method by combining the distance measure proposed by Ahmad and Dey \cite{AmirLipika2007} with the APC algorithm. Accurate clustering results are achieved with this method.
He \textit{et al.} \cite{Hemixed} extend the Squeezer algorithm \cite{He2002} which works for pure categorical datasets for clustering mixed data. In one of the versions, the numeric features are discretized to convert them to categorical features and then Squeezer algorithm is applied to the new categorical data.   
In another work, He \textit{et al.} \cite{Clusterensemble2005cs........9011H} divide the mixed data into two parts: pure numeric features and pure categorical features. A graph partitioning algorithm is used to cluster numeric data, whereas categorical data is clustered by using the Squeezer algorithm. The clustering results are combined and treated as categorical data, which is clustered by using  the Squeezer algorithm to get the final clustering results. Hai and  Susumu \cite{Hai2005} parallelize the clustering algorithm proposed by He \textit{et al.} \cite{Hemixed} to handle large datasets.

Zhao \textit{et al.} \cite{ZHAO2018264} present an ensemble method,  which creates base clustering models in sequence, for mixed dataset. The clustering models are created so that they have large diversity. The first base clustering model is created by a random partition of data points. In each run, a clustering model is generated and each data point is checked to find whether changing its cluster membership will decrease the value of a proposed optimization function. The complexity of this algorithm is quadratic. As  the start of the proposed algorithm is random, the final clustering results may be different with different initial random clusters.   

B\"{o}hm \textit{et al.} propose \cite{Bohm2010} a parameter-free clustering algorithm, INTEGRATE, for mixed data. The algorithm is based on the concept of  
MDL \cite{MDLRISSANEN1978465}. This allows the balancing of the effects of numeric and categorical features. INTEGRATE is scalable to large datasets. Behzadi \textit{et al.} \cite{parameterfree1} propose a distance hierarchy  to compute the distances for mixed datasets. A modified DBSCAN clustering algorithm is used to cluster the data and the MDL principle is used for clustering without specifying parameters.

Plant \cite{Plant:2012:DCA:2339530.2339589} propose a clustering algorithm, scale-free dependency clustering (Scenic), for mixed data. Mixed-type feature dependency patterns are detected by projecting the data points and the features into a joint low-dimensional feature space  \cite{michailidis1998}. The clusters are then obtained in the new low-dimensional embedding.

Li and Ye \cite{Li1597409} propose an incremental clustering approach for mixed data. Two different distance measures are proposed to compute the distance between clusters. In the first distance measure, separate distance measures are computed for numeric and categorical features, and then they are integrated into a new distance measure. In the second distance measure, categorical features are transformed into numeric features, and then a distance measure is computed by using all features. Similar clustering results are achieved with both  distance measures. 
Mohanavalli and Jaisakthiusing \cite{Modhaclustering} use chi-square statistics for computing the weight of each feature of mixed data. The Euclidean distance for numeric features and the Hamming distance for categorical features along with these weights are used to compute the distances. The authors did not describe about the clustering algorithm used in their paper.

Cheung and Jia \cite{CHEUNG20132228} present a general clustering framework that uses the concept of similarity between data point and cluster, and propose a unified similarity metric for mixed datasets. Accordingly, they propose an iterative clustering algorithm that finds the number of clusters automatically. Sangam and Om \cite{Sangam2015} present a sampling-based clustering algorithm for mixed datasets. The algorithm has two steps: first, a sample of data points is used for clustering, and then other points are assigned to the clusters depending upon their similarity with the clusters. They develop a hybrid similarity measure to determine the similarity between a data point and a cluster. In their method, the clustering algorithm presented by Cheung and Jia \cite{CHEUNG20132228} is used in the first step.

Lin \textit{et al.} \cite{RFclusters} present a tree-ensembles clustering algorithm, CRAFTER, for clustering high-dimensional mixed datasets. First, a random subset of data points is drawn and the random forests clustering algorithm \cite{RFC} is applied. The clustered data points are used to train tree classifiers. These trained tree-ensembles are used to cluster all of the data points.

Sangam and Om \cite{Sangam2018} present a clustering algorithm for time-evolving data streams. They propose a window-based method to detect concept drift. The data characteristics of features in the current sliding window are compared with those of the previous sliding window; the frequency is used for a categorical feature and the mean and standard deviation are used for a numeric feature. A similarity difference that exceeds the user-defined threshold indicates a concept drift.. The clustering algorithm proposed by Cheung and Jia \cite{CHEUNG20132228} is used to show the results.

Three-way clustering deals with three decisions; a data point certainly belongs to a cluster, a data point may belong to a cluster (uncertain) and a data point certainly does not  belong to a cluster. Yu \textit{et al.} \cite{threeway1} propose a three-way clustering algorithm for mixed datasets. They propose a new distance measure to compute the distance between two data points. A tree-based distance measure is proposed for categorical features. The difference between normalized feature values is used for numeric features. The algorithm uses a mixed data clustering algorithm and thresholds The references are missing from the paper, so it cannot be studied in detail. Xiong and Yu \cite{threeways2} extend this work and present an adaptive three-way clustering algorithm for mixed datasets which can produce three-way clustering without thresholds.

\section{Analysis of the Survey}
\label{sec:analysis}
The previous section reviews the majority of the key  clustering algorithms around five broad research themes for mixed data. Some of the newer and less developed areas of research are combined into the `\textit{Other}' theme. We also observed that few studies encompass more than one research theme (for example combining ideas from partitional and neural network-based clustering). However, we noted that algorithms based on partitional clustering are mostly favored by researchers and practitioners, because these algorithms are:
\begin{itemize}
    \item simpler in interpretation and implementation;
    \item linear in the number of data objects; so they scale well with big data application;
    \item easily adaptable to parallel architectures, making them more practical to apply to big data problems.
\end{itemize}

Despite these advantages, finding an appropriate similarity measure and cost function to handle mixed data remains a challenge in partitional clustering algorithms. Nonetheless, these algorithms work well in practice. The hierarchical, model-based, neural network-based, and other clustering approaches may provide better clustering outcomes; however, either they suffer from nonlinear time or space complexity or they involve making assumptions about the data distribution that may not hold in real-world scenarios. These reasons further impede progress in non-partitional clustering algorithms.

Research developments are taking place to address the problems of traditional clustering algorithms, such as the problems of cluster initialization and the number of desired clusters (for partitional algorithms) and the selection of the proper model and reasonable parameter assumptions (for model-based clustering). New trends in clustering, including subspace clustering, spectral clustering, clustering ensembles, big data clustering, and data stream clustering have been suggested for mixed datasets.

A major issue in evaluating these clustering algorithms is the choice of performance metric. In an ideal clustering scenario, class labels are not available-this is, in fact, the rationale behind performing unsupervised learning. In the absence of class labels, evaluating the performance of clustering algorithms is not straightforward. Typically, the datasets that are used to demonstrate mixed data clustering results have class labels, which are not used to perform clustering but are treated as ground truth. The final clustering results are matched with the ground truth to evaluate the performance of a clustering algorithm. Therefore, as ground truth labels are present (but are not used to perform clustering), many performance measures have been used in the literature, including F-measure, normalized mutual information, and rand index \cite{ClusteringbookJain}. However, in our survey, we found that clustering accuracy has been the most commonly used criterion for evaluating the quality of clustering results. The clustering accuracy (AC) is calculated by using the following formula:
\begin{eqnarray}
    AC = \sum\limits_{i=1}^n{c_i}/n 
\end{eqnarray}

where $c_i$ is the number of data points occurring both in the $i^{th}$ cluster
and their corresponding true class, and $n$ is the number of data points
in the dataset. The assignment of a class label to a cluster is done such that the $AC$ is maximum. 

In the literature survey, we found a lack of comparison between competitive clustering algorithms. Part of the problem is the choice of different datasets by various algorithms. It emerged that some of the popular datasets used by many researchers to evaluate their algorithms are: Heart (Cleveland), Heart (Statlog), and Australian Credit data. However, these datasets are relatively small in size, and may not be representative of real-world datasets and complex problems.

In the next section, we present several publicly available software packages for performing mixed data clustering and list some major application areas.

\section{Software and Applications}
\label{sec:public}
\subsection{Software}
As the field of mixed data clustering progresses, many researchers have made software packages and libraries available for use by the wider community. The majority of these software packages are available in R  \cite{Rcitation}.
The K-prototypes clustering algorithm \cite{Huang1998} is available in R \cite{Hunangimp}. The ClustMD package in R \cite{Clustmdimp} is the implementation of  model-based clustering for mixed data \cite{McParland}. Gower's similarity matrix \cite{Gowersimilarity} is implemented in R. The similarity matrix can be used with the partitioning around medoids tools in R or the hierarchical clustering tools to obtain final clusters \cite{MixedR}.
ClustOfVar \cite{ClustofVar} is an R package for clustering that can handle mixed datasets. 
Both a hierarchical clustering algorithm and a K-means-type partitioning algorithm are implemented in the package.
CluMix is another package in R for clustering and visualization of
mixed data \cite{CluMixR}. 
An implementation of the KAMILA \cite{Foss2016} clustering algorithm is available in R \cite{kamilaR}.
The mixed data clustering algorithm by Macbar \textit{et al.} \cite{Marbaumodelclustering} is implemented in R \cite{MacbarR}.
The Ahmad and Dey mixed data clustering algorithm \cite{AmirLipika2007} is available in Matlab \cite{AmirAhmadDeyimpl}.
A K-means-type clustering algorithm that can deal with mixed datasets is implemented in Matlab, using feature discretization \cite{MatlabMixed}. MixtComp is a C++ implementation of model-based clustering of mixed data \cite{MixtCompc++}.

\subsection{Major Application Areas} 
\label{sec:applications}
Most of the real world applications contain mixed data. Some of these application areas are (but not limited to) health, marketing, business, finance, social studies. 
Below, we present a list of major application areas where mixed-data clustering is mostly applied.

\paragraph{Health and Biology}
McParland et al \cite{McParland,McParland2016} develop mixed data clustering algorithm to study high-dimensional
numeric phenotypic data and categorical genotypic data. The study leads to a better understanding of  metabolic syndrome (MetS).
Malo \textit{et al.} \cite{Malo:2007:MDC:1420749.1420777} use mixed data clustering to study people who died of cancer between 1994 and 2006 in Hijuelas. Storlie wt al. \cite{Stoliemodel} develop model-based clustering for mixed datasets with missing feature values to cluster autism spectrum disorder.
Researchers have used various types of clustering approaches for mixed data for heart disease \cite{Modhaclustering,LiBiswasclustering,AmirLipika2007,Ahmadsubspace20111062}, occupational Medicine \cite{Saadaoui, saadaoui2015dimensionally}, digital mammograms \cite{Amirdigitalmammogram}, acute inflammations \cite{Plant:2012:DCA:2339530.2339589,Pathak2016,Ji2012129}, age of abalone snails \cite{Plant:2012:DCA:2339530.2339589}, human life span \cite{Kuri2011Clustering}, dermatology \cite{Plant:2011:IIC:2020408.2020584}, medical diagnosis \cite{Li1597409}, toxicogenomics \cite{BushelPhD}, genetic Regulation, analysis of bio-medical datasets,  \cite{BICOM} and cancer Samples Grouping \cite{Abidin2016}
\paragraph{Business and Marketing}
Hennig and Liao \cite{SocioMixed} apply mixed data clustering techniques for socio-economic stratification by using 2007 US survey data of consumer finances. Kassi \textit{et al.} \cite{Kassi7507121} develop a mixed data clustering algorithm to segment gasoline services stations in
Morocco to determine important features that can influence the profit of these service stations. Mixed data clustering has also been used in
credit approval \cite{Modhaclustering,AutomatedWeightclus,LiBiswasclustering,AmirLipika2007,Ahmadsubspace20111062}, income prediction (adult data) \cite{Hsu20081177,Modhaclustering,Jiimprovedkprototype}, marketing research \cite{morlini2010comparing}, customer behavior discovery \cite{cheng2009customer},  customer segmentation and catalog marketing \cite{Hsu200712}, customer behavior pattern discovery \cite{Cheng5366556}, motor insurance \cite{huang1997clustering} and construction management \cite{CHENG20095761}.

\paragraph{Other Applications}
Moustaki and Papageorgiou \cite{MoustakiRePEc:eee:csdana:v:48:y:2005:i:3:p:659-675} apply mixed data clustering  in archaeometry for classifying archaeological findings into groups. Philip and Ottaway \cite{Philips1983} use mixed data clustering to cluster Cypriot hooked-tang weapons. Chiodi use mixed data clustering for andrological data \cite{Chiodiclustering}.  Iam-On and Boongoen \cite{Iam-On2017} use mixed data clustering for student dropout prediction in a Thai university. Mixed data clustering has also been used in
teaching assistant evaluation \cite{LamARTclustering,Liang20122251}, class examination \cite{Hunt2WIDM:WIDM33}, petroleum recovery \cite{LamARTclustering}, intrusion detection \cite{Liu2012,Ren7603350,Li1597409}, forest cover type \cite{Mapreducemixed}, online learning systems \cite{Niuclustering}, automobiles \cite{Plant:2011:IIC:2020408.2020584}, printing process delays \cite{Barcelo-Rico2012} and country flags mining \cite{li2008weight}.

\section{Impact Areas, Challenges and Open Research Questions}

\subsection{Impact Areas}
As discussed in Section \ref{sec:applications}, mixed data clustering algorithms have been applied in various application domains. We believe that employing mixed data clustering in multiple domains is very important; however, we argue that the areas of health and business informatics will have more impact because they attempt to solve real-world problems that are related to people.

\paragraph{Health Informatics}
The majority of the data for health applications are based on either electronic health records (EHR) \cite{jensen2012mining} or sensors \cite{khan2017review}. EHR data can contain a patient's medical history, diagnoses, medications, treatment plans, immunization dates, allergies, radiology images, and laboratory and test results \cite{ehr}. EHR is a great resource to allow the deployment of evidence-based supervised and unsupervised machine learning tools to make decisions about patients' care. Therefore, EHR data is a good example of mixed data with high-impact real-world applications. Data from sensors can be either numeric (for example, motion or physiology) or categorical (for example, door open or closed). These datasets are important in building machine learning driven applications for rehabilitation, assessment of medical conditions, and detection and prediction of health-related events \cite{khan2017daad, khan2018detecting}. Application of mixed data clustering on these datasets is important in identifying medical conditions among people with disability, morbidity, and cognitive disorders. Clustering on these diverse datasets can also help in performing sex and gender based research, and  vulnerable populations and older adults.

\paragraph{Business Analytics}
Business analytic is another domain in which a large number of mixed datasets are created. Market research is an important area in this domain. Analysis of customer datasets that contain categorical features  (for example type of a customer, preference, and income group)  and numeric features (for example, age and the number of transactions) provide managers with insights about the customer behavior \cite{morlini2010comparing}. Credit card data analysis is used to predict the financial health of an individual. Generally, credit card datasets are mixed datasets. Various clustering algorithms have been applied on mixed credit datasets  \cite{Modhaclustering,AmirLipika2007}. The financial statements of a company are analyzed to assess the company's financial health; the datasets consisting of categorical features (for example, the type of the company, products and the region of the company) along with numeric features (for example, financial ratios) present better information about a company. People analytics \cite{peopleanalytics} is an emerging area: companies are interested in knowing about present and future employees to improve their productivity and satisfaction. Employee datasets consisting of categorical features (education, department, and job type) and numeric features (age, years in job, and salary) can capture information about employees better than datasets containing only one type of feature. 

\subsection{Challenges}
\label{sec:challenges}
In the previous sections, we mentioned several technical challenges for mixed-data clustering algorithms.
We now summarize those challenges for each research theme of the taxonomy with detailed ideas for future research directions .
\paragraph{Partitional Clustering}
As noted previously, one of the reasons of widespread usage of partitional clustering algorithm for mixed data is their linear time complexity with respect to the number of data points. However, the notion of center may not be clearly defined for these algorithms. Therefore, combining numeric and categorical centers to initialize these algorithms is not straightforward and it requires more research to obtain a good representation of the concept of cluster center. Another related aspect of these algorithms is finding the similarity between data objects and cluster centers. The literature suggests the development of several distance measures \cite{Huang1998,AmirLipika2007,AutomatedWeightclus}; however, the scale by which numeric and categorical distances are combined is not clear. Among the available similarity measures, there is no unanimous winner and this specific area needs more research.

The literature review suggests that cluster center initialization may help in learning consistent and robust clusters. Several methods have also been proposed for that purpose \cite{Clustercenterinitialization,Jinyin2017539,AhmadKharmonic}; however, there is no method that is both computationally inexpensive and gives consistent results in different runs. Finding good initial clusters is the key to the success of these algorithms and must be treated as an active area of research. Similarly, estimating the number of clusters in a mixed dataset is an important and challenging problem. Identifying a number of clusters that is close to the natural number of clusters in the dataset can enhance our understanding of not only the dataset but also the underlying problem.

\paragraph{Hierarchical Clustering}
The majority of hierarchical clustering algorithms rely on calculating a similarity matrix, from which clusters can be constructed. However, the similarity matrix depends on a good definition of similarity or distance. As stated above, the distance between two mixed data objects is not self-explanatory and requires more research.

\paragraph{Model-based Clustering}
As observed in the literature review, the majority of the model-based mixed data clustering algorithms suffer from high model complexity. The selection of an appropriate model is an important step in model-based clustering. There are two types of features in mixed datasets, so the selection of models for these two types of features is a challenging task. Modeling the conditional dependency between categorical and numeric features is another challenge. Selecting appropriate parametric assumptions is a difficult problem for model-based clustering. As mixed datasets have categorical features, which are not continuous, this problem is more serious for model-based clustering. As there are two types of features, identifying important features for distinguishing clusters presents a difficult task. These drawbacks may turn out be an obstacle to employing such powerful methods on large datasets to solve real-world problems. Therefore, significant effort is needed to develop models that can work with fewer parameters and offer lower model complexity.

\paragraph{Neural Network-Based Clustering}
The majority of the research work on clustering mixed data using neural networks is centered around SOM and ART. The SOM methods may lead to poor topological mappings and may not be able to match the structure of the distribution of the data \cite{du2010clustering}.
The ART models are typically governed by differential equations and have high computational complexity \cite{du2010clustering}.
There are several other areas of traditional neural network-based clustering that can be adapted for mixed datasets: for example, adaptive subspace SOM, ARTMAP, and learning vector quantization  \cite{du2010clustering}. 

\subsection{Open Research Questions and Guidelines}
In this section, we highlight several open questions that may be relevant to the different types of clustering algorithms discussed in the proposed taxonomy.

\begin{itemize}

\item Cluster ensembles have shown great promise for clustering numeric datasets by significantly improving the results of the base clustering algorithm \cite{clusteringensembles,clusteringensembles2}. 
However, more research is desirable for developing robust cluster ensemble methods for mixed datasets.

\item It is well known that real-world mixed data is imperfect;  missing values among features is one such major issue that may impair the capabilities of many existing clustering algorithms. One plausible approach is to first impute missing mixed data values \cite{audigier2016principal} and perform existing clustering methods. The other approach is to develop clustering algorithms that can handle missing data in their objective function \cite{Hunt12003429}. However, the development of, and comparison between, these two types of competitive approaches has not been investigated much and this may require attention from the research community to solve real-world problems.

\item	Various mixed datasets in application areas such as medical and socio-economics contain uncertain data because of improper data acquisition methods or inherent problems in the data acquisition. In our review, we could not find methods that can handle these types of datasets. Clustering uncertain mixed datasets is an important research direction, with applications in many domains.

\item Few researchers have developed methods for converting a mixed dataset to a pure numeric dataset, so that clustering algorithms meant for pure numeric datasets can be employed \cite{LamARTclustering,Weie17031535}. This is indeed a new perspective on the difficult problem of mixed data clustering. We further note that transformation of mixed data to numeric data does not come without loss of information. Therefore, it is an open question to the research community to develop algorithms that can reduce the adverse effects of data transformation. 

\item Clustering with deep learning approaches is an emerging area of research \cite{aljalbout2018clustering,min2018survey}. The objective (loss) function of deep learning clustering methods is primarily composed of the deep network loss and the clustering loss. Therefore, these methods differ according to the network architecture (for example, autoencoder, variational autoencoder, or generative adversarial network) or the type of clustering method (such as partitional or hierarchical). However, these methods are mostly aimed at numeric datasets; there is a great opportunity to explore mixed data clustering alongside deep learning methods.

\item As datasets increase in size and domains become complex, the majority of successful machine learning algorithms lose their interpretability and may be treated as a black box. Mixed data clustering algorithms are no exception. The idea of clustering models that are easy to explain is attractive to practitioners, such as clinicians, business analysts, geologists, and biologists. Interpretable models can assist them in making informed decisions. Unfortunately, only a few researchers have explored this area of developing interpretable mixed data clustering methods to address critical aspects of the models: for example, why a certain set of data points forms a cluster or how different clusters can be distinguished from each other \cite{plant2011inconco}. Novel research in this area will produce outcomes outside the realms of the research community. Many clustering algorithms may benefit from reducing the dimensions of multivariate mixed data, as a result of reducing their execution time and model complexity. There has been recent research in the field of feature selection for mixed data  \cite{zhang2016feature, tang2007feature}; however, combining such results with clustering has not been much explored. Selecting a subset of relevant features has the potential to enhance the interpretability of clustering algorithms as well.

\item Another repercussion of big data is ensuring the scalability of clustering algorithms, to make them useful in real-world scenarios. Parallelization of mixed data clustering algorithms is a viable approach  \cite{Mapreducemixed} to allow them to scale with increasing data size and maintain linear time complexity (especially for partitional clustering). Active research in this area is required to keep the field in synchronization with big data challenges. Similarly, developing fast and accurate online clustering algorithms to handle large streams of mixed data requires attention to address shortcomings. These include low clustering quality, evaluation of new concepts and concept drift in the underlying data, difficulties in determining cluster centers, and poor ability to deal with outliers  \cite{Mixeddatastream}. 

\item 
Subspace clustering is a viable approach to cluster large quantities of high-dimensional mixed data, though the large data problem in itself is very challenging. The extension of other subspace clustering approaches, for example, grid-based methods for mixed datasets  \cite{Parsons:2004} is key to development of clustering algorithms for high-dimensional mixed datasets.
 In subspace clustering, a data point can belong to more than one cluster and subspaces are axis-parallel \cite{subspacenew}. Research on adapting other subspace clustering approaches that have been developed for numeric datasets, such as correlation clustering  \cite{Bansal2004}, should be extended to mixed data clustering. In particular, using the correlation between numeric and categorical features to create subspaces is an innovative research area. 

\item Integration of domain knowledge into clustering is an important research area as this can improve the clustering accuracy and cluster interpretation. Constrained clustering is an approach to handling problems of this type. Constrained clustering for iterative partitional clustering methods has been proposed for mixed datasets  \cite{Wagstaff:2001:CKC:645530.655669}; however, there has been no research work on the application of constrained clustering to other approaches of clustering, such as hierarchical and density-based clustering. With the availability of large domain knowledge, there is a need to develop clustering algorithms for mixed data that can utilize this knowledge to create more accurate and interpretable clusters. 

\item	Several clustering algorithms require user-defined parameters. Therefore, the final clustering results are strongly dependent on these parameters, which include the number of desired clusters and initial clusters for iterative partitional clustering algorithms and the model selection for model-based clustering. Some efforts have been made to develop parameter-free clustering algorithms for mixed datasets \cite{Bohm2010,parameterfree1}; however, research in this field is quite open-ended.

\item Spectral clustering produces good results and does not require strong assumptions about the statistics of the clusters.
Spectral clustering  has been used to cluster mixed datasets \cite{Luo2006,Niuclustering}. The similarity matrix is the first step of spectral clustering. Each spectral clustering method for mixed datasets develops its own similarity matrix \cite{Luo2006,Niuclustering}. A large number of similarity measures are available for mixed datasets. A detailed study is required to understand which similarity measures are more useful for spectral clustering. 

\item As a pure unsupervised machine learning paradigm, true labels should not be present during clustering. Thus, evaluating the performance of clustering algorithms in this situation is not straight forward. However, in certain experimental scenarios, true labels may be present, and they can be used for matching with clustering labels. In the literature review, we also observed that \textit{accuracy} has been reported by many researchers as a performance metric for clustering algorithms (see Section \ref{sec:analysis}). A major problem with using \textit{accuracy} or other confusion matrix-based performance measures is that they assume a direct correspondence between true and clustering labels. However, clustering labels are arbitrary and matching them with true labels is non-trivial. With small data size and a small number of natural clusters, this technique of matching true and clustering labels may be feasible, with support from domain knowledge. However, \textit{accuracy} will be difficult to comprehend \textit{accuracy} as the number of clusters and data points increase. Therefore, in experimental scenarios where the true labels are known, performance metrics such as adjusted rand index, normalized mutual information, homogeneity, completeness, and the V-measure \cite{Rendon:2011:CIE:1959666.1959695} are more relevant and should be widely adopted. For real-world clustering problems where the true labels are not be present, performance indexes such as the silhouette coefficient, Calinski-Harabaz Index, and Davies-Bouldin Index  \cite{Internalindices} should be used.

\item	A ubiquitous problem that has been highlighted in this literature review is that the majority of clustering algorithms tested their methods on a few publicly available datasets. Moreover, several researchers showed results on datasets that were not available to the wider community. We believe that creating a community-based mixed data repository not only provides opportunities to compare existing clustering methods and set benchmarks but also encourages the development of new algorithms at a faster pace. Furthermore, we believe that sharing and contributing clustering algorithms' code in the public domain, by means such as R packages, Python libraries, and Java classes is useful for quickly comparing and testing existing and new methods. As discussed in Section  \ref{sec:public}, some software packages have been made public; more effort will certainly benefit the research community.

\end{itemize}

In this paper, we identified five major research themes for the study of mixed data clustering and presented a comprehensive state-of-the-art survey of literature within them. We discussed the challenges and future directions within each research theme, and discussed several high-impact application areas, open research questions, and guidelines for making progress in the field. This survey paper should guide researchers to develop an in-depth understanding of the field of mixed data clustering and help generate new ideas to make significant contributions to solve real-world problems.

\bibliography{mixeddataref}

\begin{thebibliography}{100}
\providecommand{\url}[1]{#1}
\csname url@samestyle\endcsname
\providecommand{\newblock}{\relax}
\providecommand{\bibinfo}[2]{#2}
\providecommand{\BIBentrySTDinterwordspacing}{\spaceskip=0pt\relax}
\providecommand{\BIBentryALTinterwordstretchfactor}{4}
\providecommand{\BIBentryALTinterwordspacing}{\spaceskip=\fontdimen2\font plus
\BIBentryALTinterwordstretchfactor\fontdimen3\font minus
  \fontdimen4\font\relax}
\providecommand{\BIBforeignlanguage}[2]{{%
\expandafter\ifx\csname l@#1\endcsname\relax
\typeout{** WARNING: IEEEtran.bst: No hyphenation pattern has been}%
\typeout{** loaded for the language `#1'. Using the pattern for}%
\typeout{** the default language instead.}%
\else
\language=\csname l@#1\endcsname
\fi
#2}}
\providecommand{\BIBdecl}{\relax}
\BIBdecl

\bibitem{kmeanclusterinitialization1}
S.~S. Khan and A.~Ahmad, ``Cluster center initialization algorithm for k-means
  clustering,'' \emph{Pattern Recognition Letters}, vol.~25, no.~11, pp.
  1293--1302, 2004.

\bibitem{ClusteringbookJain}
A.~Jain and R.~Dubes, \emph{Algorithms for Clustering Data}.\hskip 1em plus
  0.5em minus 0.4em\relax Prentice Hall, 1988.

\bibitem{Booksmachine}
C.~M. Bishop, \emph{Pattern {R}ecognition and {M}achine {L}earning}.\hskip 1em
  plus 0.5em minus 0.4em\relax Springer-Verlag New York Inc, 2008.

\bibitem{WEKAwitten1}
I.~H. Witten and E.~Frank, \emph{Data {M}ining: {P}ractical {M}achine
  {L}earning {T}ools and {T}echniques.}, 2nd~ed.\hskip 1em plus 0.5em minus
  0.4em\relax Morgan Kaufmann San Francisco, CA, 2005.

\bibitem{Categoricalsimilarity}
S.~Boriah, V.~Chandola, and V.~Kumar, ``Similarity measures for categorical
  data: A comparative evaluation,'' in \emph{Proceedings of the 2008 SIAM
  International Conference on Data Mining}, 2008, pp. 243--254.

\bibitem{AmirLipika2007}
A.~Ahmad and L.~Dey, ``A k-mean clustering algorithm for mixed numeric and
  categorical data,'' \emph{Data and Knowledge Engineering}, vol.~63, no.~2,
  pp. 503--527, 2007.

\bibitem{SocioMixed}
C.~Hennig and T.~F. Liao, ``How to find an appropriate clustering for
  mixed-type variables with application to socio-economic stratification,''
  \emph{Journal of the Royal Statistical Society Series C}, vol.~62, no.~3, pp.
  309--369, 2013.

\bibitem{Morlini2010}
I.~I.~Morlini and S.~Zani, \emph{Comparing Approaches for Clustering Mixed Mode
  Data: An Application in Marketing Research}.\hskip 1em plus 0.5em minus
  0.4em\relax Springeg, 2010, pp. 49--57.

\bibitem{Huang1998}
Z.~Huang, ``Extensions to the k-means algorithm for clustering large data sets
  with categorical values,'' \emph{Data Min. Knowl. Discov.}, vol.~2, no.~3,
  pp. 283--304, 1998.

\bibitem{Huang1997}
------, ``Clustering large data sets with mixed numeric and categorical
  values,'' in \emph{Proceedings of the $1^{st}$ Pacific Asia Knowledge
  Discovery and Data Mining Conference}.\hskip 1em plus 0.5em minus 0.4em\relax
  Singapore: World Scientific, 1997, pp. 21--34.

\bibitem{Reviewdistance}
M.~V. Velden, A.~I. D'enza, and A.~Markos, ``Distance‐based clustering of
  mixed data,'' \emph{WIREs Computational Statistics}, 2018.

\bibitem{Reviewmodelbased}
A.~H. Foss, M.~Markatou, and B.~Ray, ``Distance metrics and clustering methods
  for mixed‐type data,'' \emph{International Statistical Review}, 2018.

\bibitem{mixedddatareview}
K.~Balaji and K.~Lavanya, ``Clustering algorithms for mixed datasets: A
  review,'' \emph{International Journal of Pure and Applied Mathematics},
  vol.~18, no.~7, pp. 547--556, 2018.

\bibitem{miyamotoreview}
S.~Miyamoto, V.~Huynh, and S.~Fujiwara, ``Methods for clustering categorical
  and mixed data: An overview and new algorithms,'' in \emph{Integrated
  Uncertainty in Knowledge Modelling and Decision Making}, 2018, pp. 75--86.

\bibitem{AutomatedWeightclus}
J.~Z. Huang, M.~K. Ng, H.~Rong, and Z.~Li, ``Automated variable weighting in
  k-means type clustering,'' \emph{IEEE Transactions on Pattern Analysis and
  Machine Intelligence}, vol.~27, no.~5, pp. 657--668, 2005.

\bibitem{Modhaclustering}
D.~S. Modha and W.~S. Spangler, ``Feature weighting in k-means clustering,''
  \emph{Machine Learning}, vol.~52, no.~3, pp. 217--237, Sep. 2003.

\bibitem{Mixeddatastream}
J.~Chen and H.~He, ``A fast density-based data stream clustering algorithm with
  cluster centers self-determined for mixed data,'' \emph{Information
  Sciences}, vol. 345, pp. 271--293, 2016.

\bibitem{Ren7603350}
M.~Ren, P.~Liu, Z.~Wang, and X.~Pan, ``An improved mixed-type data based kernel
  clustering algorithm,'' in \emph{2016 12th International Conference on
  Natural Computation, Fuzzy Systems and Knowledge Discovery (ICNC-FSKD)},
  2016, pp. 1205--1209.

\bibitem{Jiimprovedkprototype}
J.~Ji, T.~Bai, C.~Zhou, C.~Ma, and Z.~Wang, ``An improved k-prototypes
  clustering algorithm for mixed
  numerihttps://www.overleaf.com/project/5c177f6e8c0fb95e14cd9a36c and
  categorical data,'' \emph{Neurocomputing}, vol. 120, pp. 590 -- 596, 2013.

\bibitem{Sangam2018old}
R.~K. Sangam, , and H.~Om, ``An equi-biased k-prototypes algorithm for
  clustering mixed-type data,'' \emph{S{\={a}}dhan{\={a}}}, vol.~43, no.~3,
  p.~37, 2018.

\bibitem{Geneticclustering1}
D.~K. Roy and L.~K. Sharma, ``Genetic k-means clustering algorithm for mixed
  numeric and categorical data sets,'' \emph{International Journal of
  Artificial Intelligence}, vol.~1, no.~2, p. 23�28, 2010.

\bibitem{Wang:2015}
C.~Wang, C.~Chi, W.~Zhou, and R.~Wong, ``Coupled interdependent attribute
  analysis on mixed data,'' in \emph{Proceedings of the Twenty-Ninth AAAI
  Conference on Artificial Intelligence}, ser. AAAI'15, 2015, pp. 1861--1867.

\bibitem{Weie17031535}
M.~Wei, T.~W.~S. Chow, and R.~H.~M. Chan, ``Clustering heterogeneous data with
  k-means by mutual information-based unsupervised feature transformation,''
  \emph{Entropy}, vol.~17, no.~3, pp. 1535--1548, 2015.

\bibitem{Zhaoclsuering2007}
W.~Zhao, W.~Dai, and C.~Tang, ``K-centers algorithm for clustering mixed type
  data,'' in \emph{Proceedings of the $11^{th}$ Pacific-Asia Conference on
  Advances in Knowledge Discovery and Data Mining}, ser. PAKDD'07, 2007, pp.
  1140--1147.

\bibitem{Chiodiclustering}
M.~Chiodi, ``A partition type method for clustering mixed data,'' \emph{Rivista
  di Statistica Applicata}, vol.~2, p. 135�147, 1990.

\bibitem{Mapreducemixed}
M.~A.~B. Kacem, C.~E.~B. N'cir, and N.~Essoussi, ``Mapreduce-based k-prototypes
  clustering method for big data,'' in \emph{2015 IEEE International Conference
  on Data Science and Advanced Analytics (DSAA)}, 2015, pp. 1--7.

\bibitem{GridClustering}
H.~Jang, B.~Kim, J.~Kim, and S.~Jung, ``An efficient grid-based k-prototypes
  algorithm for sustainable decision-making on spatial objects,''
  \emph{Sustainability}, vol.~10, no.~8, 2018.

\bibitem{Barcelo-Rico2012}
F.~Barcelo-Rico and D.~Jose-Luis, ``Geometrical codification for clustering
  mixed categorical and numerical databases,'' \emph{Journal of Intelligent
  Information Systems}, vol.~39, no.~1, pp. 167--185, 2012.

\bibitem{CHENG20095761}
Y.~Cheng and S.~Leu, ``Constraint-based clustering and its applications in
  construction management,'' \emph{Expert Systems with Applications}, vol.~36,
  no. 3, Part 2, pp. 5761 -- 5767, 2009.

\bibitem{Ahmad2005}
A.~Ahmad and L.~Dey, \emph{Algorithm for Fuzzy Clustering of Mixed Data with
  Numeric and Categorical Attributes}.\hskip 1em plus 0.5em minus 0.4em\relax
  Berlin, Heidelberg: Springer Berlin Heidelberg, 2005, pp. 561--572.

\bibitem{Ji2012129}
J.~Ji, W.~Pang, C.~Zhou, X.~Han, and Z.~Wang, ``A fuzzy k-prototype clustering
  algorithm for mixed numeric and categorical data,'' \emph{Knowledge-Based
  Systems}, vol.~30, pp. 129--135, 2012.

\bibitem{Kuri-Morales:2011:CHT:2178145.2178169}
A.~Kuri-Morales, D.~Trejo-Ba\~{n}os, and L.~E. Cortes-Berrueco, ``Clustering of
  heterogeneously typed data with soft computing - a case study,'' in
  \emph{Proceedings of the 10th International Conference on Artificial
  Intelligence: Advances in Soft Computing - Volume Part II}.\hskip 1em plus
  0.5em minus 0.4em\relax Springer-Verlag, 2011, pp. 235--248.

\bibitem{Clustercenterinitialization}
J.~Ji, W.~Pang, Y.~Zheng, Z.~Wang, Z.~Ma, and L.~Zhang, ``A novel cluster
  center initialization method for the k-prototypes algorithms using centrality
  and distance,'' \emph{Applied Mathematics and Information Sciences}, vol.~9,
  no.~6, pp. 2933--2942, 2015.

\bibitem{Jinyin2017539}
J.~Chen, X.~L. Xiang, H.~Zheng, and X.~Bao, ``A novel cluster center fast
  determination clustering algorithm,'' \emph{Applied Soft Computing}, vol.~57,
  pp. 539--555, 2017.

\bibitem{WANGCHAMHAN}
T.~Wangchamhan, S.~Chiewchanwattana, and K.~Sunat, ``Efficient algorithms based
  on the k-means and chaotic league championship algorithm for numeric,
  categorical, and mixed-type data clustering,'' \emph{Expert Systems with
  Applications}, vol.~90, pp. 146--167, 2017.

\bibitem{crow}
K.~Lakshmi, N.~V. Karthikeyani, S.~Shanthi, and S.~Parvathavarthini,
  ``Clustering mixed datasets using k-prototype algorithm based on crow-search
  optimization,'' in \emph{Developments and Trends in Intelligent Technologies
  and Smart Systems}, 2017, pp. 129--150.

\bibitem{AhmadKharmonic}
A.~A.~S. Hashmi, ``K-harmonic means type clustering algorithm for mixed
  datasets,'' \emph{Applied Soft Computing}, vol.~48, no.~C, pp. 39--49, 2016.

\bibitem{Liang20122251}
J.~Liang, X.~Zhao, D.~Li, F.~Cao, and C.~Dang, ``Determining the number of
  clusters using information entropy for mixed data,'' \emph{Pattern
  Recognition}, vol.~45, no.~6, pp. 2251--2265, 2012.

\bibitem{YAO2018166}
X.~Yao, S.~Ge, H.~Kong, and H.~Ning, ``An improved clustering algorithm and its
  application in wechat sports users analysis,'' \emph{Procedia Computer
  Science}, vol. 129, pp. 166 -- 174, 2018, 2017 International conference on
  identification, information and knowledge in the internet of things.

\bibitem{Philips1983}
G.~Philip and B.~S. Ottaway, ``Mixed data cluster analysis: an illustration
  using cypriot hooked-tang weapons,'' \emph{Archaeometry}, vol.~25, no.~2, pp.
  119--133, 1983.

\bibitem{LiBiswasclustering}
C.~Li and G.~Biswas, ``Unsupervised learning with mixed numeric and nominal
  data,'' \emph{IEEE Transaction on Knowledge and Data Engineering}, vol.~14,
  no.~4, pp. 673--690, 2002.

\bibitem{Chiu:2001:RSC:502512.502549}
T.~C. D.~P. Fang, J.~Chen, Y.~Wang, and C.~Jeris, ``A robust and scalable
  clustering algorithm for mixed type attributes in large database
  environment,'' in \emph{Proceedings of the $7^{th}$ ACM SIGKDD International
  Conference on Knowledge Discovery and Data Mining}, ser. KDD '01, 2001, pp.
  263--268.

\bibitem{HsuDistanceHierarchy20074474}
C.~C. Hsu, C.~G. Chen, and Y.~Su, ``Hierarchical clustering of mixed data based
  on distance hierarchy,'' \emph{Information Sciences}, vol. 177, no.~20, pp.
  4474--4492, 2007.

\bibitem{Hsu200712}
C.~Hsu and Y.~Chen, ``Mining of mixed data with application to catalog
  marketing,'' \emph{Expert Systems with Applications}, vol.~32, no.~1, pp. 12
  -- 23, 2007.

\bibitem{Hsu20081177}
C.~C. Hsu and Y.~P. Huang, ``Incremental clustering of mixed data based on
  distance hierarchy,'' \emph{Expert Systems with Applications}, vol.~35,
  no.~3, pp. 1177 -- 1185, 2008.

\bibitem{Shih2010}
M.~Shih, J.~Jheng, and L.~Lai, ``A two-step method for clustering mixed
  categroical and numeric data,'' \emph{Tamkang Journal of Science and
  Engineering}, vol.~13, no.~1, pp. 11--19, 2010.

\bibitem{Lim}
J.~Lim, J.~Jun, S.~Kim, and D.~McLeod, ``A framework for clustering mixed
  attribute type datasets,'' in \emph{Proc. of the 4th Int. Con. on Emerging
  Databases (EDB 2012)}, 2012.

\bibitem{Chaesimilarity}
S.~S. K~J.~Chae and W.~Y. Yang, ``Cluster analysis with balancing weight on
  mixed-type data,'' \emph{The Korean Communications in Statistics}, vol.~13,
  no.~3, pp. 719--732, 2006.

\bibitem{Autoclass}
P.~Cheeseman and J.~Stutz, \emph{Advances in Knowledge Discovery and Data
  Mining}.\hskip 1em plus 0.5em minus 0.4em\relax AAAI Press/MIT Press, Menlo
  Park CA, 1996, ch. Bayesian Classification (AutoClass): Theory and Results,
  pp. 153--180.

\bibitem{Everittmixedmodel}
B.~S. Everitt, ``A finite mixture model for the clustering of mixed-mode
  data,'' \emph{Statistics and Probability Letters}, vol.~6, no.~5, pp.
  305--309, 1988.

\bibitem{MoustakiRePEc:eee:csdana:v:48:y:2005:i:3:p:659-675}
I.~Moustaki and I.~Papageorgiou, ``Latent class models for mixed variables with
  applications in archaeometry,'' \emph{Computational Statistics and Data
  Analysis}, vol.~48, no.~3, pp. 659--675, 2005.

\bibitem{Browne20122976}
R.~P. Browne and P.~D. McNicholas, ``Model-based clustering, classification,
  and discriminant analysis of data with mixed type,'' \emph{Journal of
  Statistical Planning and Inference}, vol. 142, no.~11, pp. 2976--2984, 2012.

\bibitem{BICOM}
B.~Andreopoulos, A.~An, and X.~Wang, ``Bi-level clustering of mixed categorical
  and numerical biomedical data,'' \emph{{IJDMB}}, vol.~1, no.~1, pp. 19--56,
  2006.

\bibitem{Hunt3}
L.~Hunt and M.~Jorgensen, ``Mixture model clustering of data sets with
  categorical and continuous variables,'' in \emph{Information, Statistics and
  Induction in Science}.\hskip 1em plus 0.5em minus 0.4em\relax Singapore:
  World Scientific, 1996, pp. 375--384.

\bibitem{Lawrence1996}
C.~J. Lawrence and W.~J. Krzanowski, ``Mixture separation for mixed-mode
  data,'' \emph{Statistics and Computing}, vol.~6, no.~1, pp. 85--92, 1996.

\bibitem{McParland2016}
D.~McParland and I.~C. Gormley, ``Model based clustering for mixed data:
  clustmd,'' \emph{Adv. Data Analysis and Classification}, vol.~10, no.~2, pp.
  155--169, 2016.

\bibitem{Saadaoui}
F.~Sa{\^a}daoui, P.~R. Bertrand, G.~Boudet, K.~Rouffiac, F.~Dutheil, and
  A.~Chamoux, ``A dimensionally reduced clustering methodology for
  heterogeneous occupational medicine data mining,'' \emph{IEEE Transactions on
  NanoBioscience}, vol.~14, no.~7, pp. 707--715, 2015.

\bibitem{McParland}
C.~P. D.~McParland, L.~Brennan, H.~Roche, and I.~C. Gormley, ``Clustering high
  dimensional mixed data to uncover sub-phenotypes: joint analysis of
  phenotypic and genotypic data,'' \emph{CoRR}, vol. abs/1604.01686, 2016.

\bibitem{Rajan:2016}
V.~Rajan and S.~Bhattacharya, ``Dependency clustering of mixed data with
  gaussian mixture copulas,'' in \emph{Proceedings of the Twenty-Fifth
  International Joint Conference on Artificial Intelligence}, ser. IJCAI'16,
  2016, pp. 1967--1973.

\bibitem{Tekumalla2017}
L.~S. Tekumalla, V.~V.~Rajan, and C.~Bhattacharyya, ``Vine copulas for mixed
  data : multi-view clustering for mixed data beyond meta-gaussian
  dependencies,'' \emph{Machine Learning}, vol. 106, no.~9, pp. 1331--1357,
  2017.

\bibitem{Marbaumodelclustering}
M.~Marbac, C.~Biernacki, and V.~Vandewalle, ``Model-based clustering of
  gaussian copulas for mixed data,'' \emph{Communications in Statistics -
  Theory and Methods}, vol.~46, no.~23, pp. 11\,635--11\,656, 2017.

\bibitem{Foss2016}
A.~Foss, M.~Markatou, B.~Ray, and A.~Heching, ``A semiparametric method for
  clustering mixed data,'' \emph{Machine Learning}, vol. 105, no.~3, pp.
  419--458, 2016.

\bibitem{DoringFuzzy1336254}
C.~Doring, C.~Borgelt, and R.~Kruse, ``Fuzzy clustering of quantitative and
  qualitative data,'' in \emph{Fuzzy Information, 2004. Processing NAFIPS '04.
  IEEE Annual Meeting of the}, vol.~1, 2004, pp. 84--89 Vol.1.

\bibitem{Chatzis20118684}
S.~P. Chatzis, ``A fuzzy c-means-type algorithm for clustering of data with
  mixed numeric and categorical attributes employing a probabilistic
  dissimilarity functional,'' \emph{Expert Systems with Applications}, vol.~38,
  no.~7, pp. 8684--8689, 2011.

\bibitem{Pathak2016}
A.~Pathak and N.~R. Pal, ``Clustering of mixed data by integrating fuzzy,
  probabilistic, and collaborative clustering framework,'' \emph{International
  Journal of Fuzzy Systems}, vol.~18, no.~3, pp. 339--348, 2016.

\bibitem{SOM1}
H.~P. Devaraj and M.~Punithavalli, ``An integrated framework for mixed data
  clustering using self organizing map,'' \emph{Journal of Computer Science},
  vol.~7, no.~11, pp. 1639--1645, 2011.

\bibitem{Hsu:2006:GSM:2325818.2327595}
C.~Hsu, ``Generalizing self-organizing map for categorical data,'' \emph{IEEE
  Transactions on Neural Networks}, vol.~17, no.~2, pp. 294--304, 2006.

\bibitem{GViSOM}
C.~Hsu and S.~Lin, ``Visualized analysis of multivariate mixed-type data via an
  extended self-organizing map,'' in \emph{The 6th International Conference on
  Information Technology and Applications (ICITA 2009)}, 2006, pp. 218--223.

\bibitem{GsomIeee}
------, ``Visualized analysis of mixed numeric and categorical data via
  extended self-organizing map,'' \emph{IEEE Transactions on Neural Networks
  and Learning Systems}, vol.~23, no.~1, pp. 72--86, 2012.

\bibitem{TAI20122856}
W.~Tai and C.~Hsu, ``Growing self-organizing map with cross insert for
  mixed-type data clustering,'' \emph{Applied Soft Computing}, vol.~12, no.~9,
  pp. 2856 -- 2866, 2012.

\bibitem{chenning}
N.~N.~Chen and N.~C. Marques, ``An extension of self-organizing maps to
  categorical data,'' in \emph{Progress in Artificial Intelligence}.\hskip 1em
  plus 0.5em minus 0.4em\relax Berlin, Heidelberg: Springer Berlin Heidelberg,
  2005, pp. 304--313.

\bibitem{DELCOSO2015246}
C.~del Coso, D.~Fustes, C.~Dafonte, F.~J. Novoa, J.~M. Rodriguez-Pedreira, and
  B.~Arcay, ``Mixing numerical and categorical data in a self-organizing map by
  means of frequency neurons,'' \emph{Applied Soft Computing}, vol.~36, pp. 246
  -- 254, 2015.

\bibitem{Noorbehbahani2015}
F.~Noorbehbahani, S.~R. Mousavi, and A.~Mirzaei, ``An incremental mixed data
  clustering method using a new distance measure,'' \emph{Soft Computing},
  vol.~19, no.~3, pp. 731--743, 2015.

\bibitem{LamARTclustering}
D.~Lam, M.~Wei, and D.~Wunsch, ``Clustering data of mixed categorical and
  numerical type with unsupervised feature learning,'' \emph{IEEE Access},
  vol.~3, pp. 1605--1613, 2015.

\bibitem{Luo2006}
H.~Luo, F.~Kong, and Y.~Li, \emph{Clustering Mixed Data Based on Evidence
  Accumulation}.\hskip 1em plus 0.5em minus 0.4em\relax Springer Berlin
  Heidelberg, 2006, pp. 348--355.

\bibitem{DAVID2012416}
G.~David and A.~Averbuch, ``Spectralcat: Categorical spectral clustering of
  numerical and nominal data,'' \emph{Pattern Recognition}, vol.~45, no.~1, pp.
  416 -- 433, 2012.

\bibitem{Niuclustering}
K.~Niu, Z.~Niu, Y.~Su, C.~Wang, H.~Lu, and J.~Guan, ``A coupled user clustering
  algorithm based on mixed data for web-based learning systems,''
  \emph{Mathematical Problems in Engineering}, vol. 2015, 2015.

\bibitem{Ahmadsubspace20111062}
A.~Ahmad and L.~Dey, ``A k-means type clustering algorithm for subspace
  clustering of mixed numeric and categorical datasets,'' \emph{Pattern
  Recognition Letters}, vol.~32, no.~7, pp. 1062--1069, 2011.

\bibitem{Subspaceclustering2}
H.~Jia and Y.~M. Cheung, ``Subspace clustering of categorical and numerical
  data with an unknown number of clusters,'' \emph{IEEE Transactions on Neural
  Networks and Learning Systems}, vol.~PP, no.~99, pp. 1--18, 2017.

\bibitem{Plant:2011:IIC:2020408.2020584}
C.~Plant and C.~B\"{o}hm, ``Inconco: Interpretable clustering of numerical and
  categorical objects,'' in \emph{Proceedings of the 17th ACM SIGKDD
  International Conference on Knowledge Discovery and Data Mining}, ser. KDD
  '11, 2011, pp. 1127--1135.

\bibitem{DU201746}
M.~Du, S.~Ding, and Y.~Xue, ``A novel density peaks clustering algorithm for
  mixed data,'' \emph{Pattern Recognition Letters}, vol.~97, pp. 46 -- 53,
  2017.

\bibitem{DING2017294}
S.~Ding, M.~Du, T.~Sun, X.~Xu, and Y.~Xue, ``An entropy-based density peaks
  clustering algorithm for mixed type data employing fuzzy neighborhood,''
  \emph{Knowledge-Based Systems}, vol. 133, pp. 294 -- 313, 2017.

\bibitem{Liu2017}
X.~Liu, Q.~Yang, and L.~He, ``A novel dbscan with entropy and probability for
  mixed data,'' \emph{Cluster Computing}, vol.~20, no.~2, pp. 1313--1323, Jun
  2017.

\bibitem{OclusteringOracle}
M.~C.~B. Milenova, ``Clustering large databases with numeric and nominal values
  using orthogonal projections,'' Oracle Data Mining Technologies, Oracle
  Corporation, Tech. Rep., 2002.

\bibitem{cobweb3:a}
K.~Mckusick and K.~Thompson, ``Cobweb/3: A portable implementation,'' NASA Ames
  Research Center, Tech. Rep. FIA-90-6-18-2, 1990.

\bibitem{Reich:1991:FUA:120441.120453}
Y.~Reich and J.~S. Fenves, ``The formation and use of abstract concepts in
  design,'' in \emph{Concept Formation Knowledge and Experience in Unsupervised
  Learning}, D.~H. Fisher, Jr., M.~J. Pazzani, and P.~Langley, Eds.\hskip 1em
  plus 0.5em minus 0.4em\relax San Francisco, CA, USA: Morgan Kaufmann
  Publishers Inc., 1991, pp. 323--353.

\bibitem{DiCiaccio2001}
A.~D. Ciaccio, \emph{MIXISO: a Non-Hierarchical Clustering Method for
  Mixed-Mode Data}.\hskip 1em plus 0.5em minus 0.4em\relax Berlin, Heidelberg:
  Springer Berlin Heidelberg, 2001, pp. 27--34.

\bibitem{Sowjanya}
A.~M. Sowjanya and M.~Shashi, ``A cluster feature-based incremental clustering
  approach to mixed data,'' \emph{Journal of Computer Science}, vol.~7, no.~12,
  pp. 1875--1880, 2011.

\bibitem{Frey07clusteringby}
B.~J. Frey and D.~Dueck, ``Clustering by passing messages between data
  points,'' \emph{Science}, vol. 315, p. 2007, 2007.

\bibitem{Affinityclustering}
K.~Zhang and X.~Gu, ``An affinity propagation clustering algorithm for mixed
  numeric and categorical datasets,'' \emph{Mathematical Problems in
  Engineering}, 2014.

\bibitem{Hemixed}
Z.~He, X.~Xu, and S.~Deng, ``Scalable algorithms for clustering large datasets
  with mixed type attributes,'' \emph{International Journal of Intelligent
  Systems}, vol.~20, no.~10, pp. 1077--1089, 2005.

\bibitem{Clusterensemble2005cs........9011H}
------, ``{Clustering Mixed Numeric and Categorical Data: A Cluster Ensemble
  Approach},'' \emph{eprint arXiv:cs/0509011}, 2005.

\bibitem{Hai2005}
N.~T.~M. Hai and H.~Susumu, \emph{Performances of Parallel Clustering Algorithm
  for Categorical and Mixed Data}.\hskip 1em plus 0.5em minus 0.4em\relax
  Springer Berlin Heidelberg, 2005, pp. 252--256.

\bibitem{ZHAO2018264}
X.~Zhao, F.~Cao, and J.~Liang, ``A sequential ensemble clustering generation
  algorithm for mixed data,'' \emph{Applied Mathematics and Computation}, vol.
  335, pp. 264 -- 277, 2018.

\bibitem{Bohm2010}
C.~B{\"o}hm, S.~Goebl, A.~Oswald, C.~Plant, M.~Plavinski, and
  B.~Wackersreuther, ``Integrative parameter-free clustering of data with mixed
  type attributes,'' in \emph{Pacific-asia conference on knowledge discovery
  and data mining}.\hskip 1em plus 0.5em minus 0.4em\relax Springer, 2010, pp.
  38--47.

\bibitem{parameterfree1}
S.~Behzadi, Sahar, M.~A. Ibrahim, and C.~Plant, ``Parameter free mixed-type
  density-based clustering,'' in \emph{Database and Expert Systems
  Applications}, 2018, pp. 19--34.

\bibitem{Plant:2012:DCA:2339530.2339589}
C.~Plant, ``Dependency clustering across measurement scales,'' in
  \emph{Proceedings of the 18th ACM SIGKDD International Conference on
  Knowledge Discovery and Data Mining}, ser. KDD '12, 2012, pp. 361--369.

\bibitem{Li1597409}
X.~Li and N.~Ye, ``A supervised clustering and classification algorithm for
  mining data with mixed variables,'' \emph{IEEE Transactions on Systems, Man,
  and Cybernetics - Part A: Systems and Humans}, vol.~36, no.~2, pp. 396--406,
  2006.

\bibitem{CHEUNG20132228}
Y.~Cheung and H.~Jia, ``Categorical-and-numerical-attribute data clustering
  based on a unified similarity metric without knowing cluster number,''
  \emph{Pattern Recognition}, vol.~46, no.~8, pp. 2228 -- 2238, 2013.

\bibitem{Sangam2015}
R.~S. Sangam and H.~Om, ``Hybrid data labeling algorithm for clustering large
  mixed type data,'' \emph{Journal of Intelligent Information Systems},
  vol.~45, no.~2, pp. 273--293, 2015.

\bibitem{RFclusters}
S.~Lin, B.~Azarnoush, and G.~Runger, ``Crafter: a tree-ensemble clustering
  algorithm for static datasets with mixed attributes and high
  dimensionality,'' \emph{IEEE Transactions on Knowledge and Data Engineering},
  vol.~30, no.~9, pp. 1686--1696, 2018.

\bibitem{Sangam2018}
R.~S. Sangam and H.~Om, ``Equi-clustream: a framework for clustering time
  evolving mixed data,'' \emph{Advances in Data Analysis and Classification},
  vol.~12, no.~4, pp. 973--995, Dec 2018.

\bibitem{threeway1}
H.~Yu, Z.~Chang, and B.~Zhou, ``A novel three-way clustering algorithm for
  mixed-type data,'' in \emph{2017 IEEE International Conference on Big
  Knowledge (ICBK)}, 2017, pp. 119--126.

\bibitem{MacQueen1967}
J.~B. MacQueen, ``Some methods for classification and analysis of multivariate
  observations,'' in \emph{Proc. of the fifth Berkeley Symposium on
  Mathematical Statistics and Probability}, L.~M.~L. Cam and J.~Neyman, Eds.,
  vol.~1.\hskip 1em plus 0.5em minus 0.4em\relax University of California
  Press, 1967, pp. 281--297.

\bibitem{khan2007computation}
S.~S. Khan and S.~Kant, ``Computation of initial modes for k-modes clustering
  algorithm using evidence accumulation,'' in \emph{Proceedings of the 20th
  international joint conference on Artifical intelligence}.\hskip 1em plus
  0.5em minus 0.4em\relax Morgan Kaufmann Publishers Inc., 2007, pp.
  2784--2789.

\bibitem{kModeclusterinitilization1}
S.~S. Khan and A.~Ahmad, ``Cluster center initialization algorithm for k-modes
  clustering,'' \emph{Expert Syst. Appl.}, vol.~40, no.~18, pp. 7444--7456,
  2013.

\bibitem{Lu}
Y.~Lu, S.~Lu, F.~Fotouhi, Y.~Deng, and J.~S. Brown, ``Fgka: A fast genetic
  k-means clustering algorithm,'' in \emph{Proceedings of the 2004 ACM
  Symposium on Applied Computing}.\hskip 1em plus 0.5em minus 0.4em\relax New
  York, NY, USA: ACM, 2004, pp. 622--623.

\bibitem{Dean:2008:MSD:1327452.1327492}
J.~Dean and S.~Ghemawat, ``Mapreduce: Simplified data processing on large
  clusters,'' \emph{Communications of the ACM}, vol.~51, no.~1, pp. 107--113,
  2008.

\bibitem{Wagstaff:2001:CKC:645530.655669}
K.~Wagstaff, C.~Cardie, S.~Rogers, and S.~Schr\"{o}dl, ``Constrained k-means
  clustering with background knowledge,'' in \emph{Proceedings of the $18^{th}$
  Conference on Machine Learning}, ser. ICML '01, 2001, pp. 577--584.

\bibitem{fuzzyYANG19931}
M.~S. Yang, ``A survey of fuzzy clustering,'' \emph{Mathematical and Computer
  Modelling}, vol.~18, no.~11, pp. 1--16, 1993.

\bibitem{El-Sonbaty}
Y.~El-Sonbaty and M.~A. Ismail, ``Fuzzy clustering for symbolic data,''
  \emph{IEEE Transactions on Fuzzy Systems}, vol.~6, no.~2, pp. 195--204, 1998.

\bibitem{fuzzycmenas1}
J.~C. Dunn, ``A fuzzy relative of the isodata process and its use in detecting
  compact well-separated clusters,'' \emph{Journal of Cybernetics}, vol.~3,
  no.~3, pp. 32--57, 1973.

\bibitem{fuzycmenas2}
J.~C. Bezdek, \emph{Pattern Recognition with Fuzzy Objective Function
  Algorithms}.\hskip 1em plus 0.5em minus 0.4em\relax Springer US, 1981, ch.
  Pattern Recognition with Fuzzy Objective Function.

\bibitem{khan2003computing}
S.~S. Khan and A.~Ahmad, ``Computing initial points using density based
  multiscale data condensation for clustering categorical data,'' in
  \emph{$2^{nd}$ International Conference on Applied Artificial Intelligence,
  ICAAI}, 2003.

\bibitem{Rodriguez1492}
A.~Rodriguez and A.~Laio, ``Clustering by fast search and find of density
  peaks,'' \emph{Science}, vol. 344, no. 6191, pp. 1492--1496, 2014.

\bibitem{LeagueChampionshipAlgorithm}
A.~H. Kashan, ``League championship algorithm: A new algorithm for numerical
  function optimization,'' in \emph{2009 International Conference of Soft
  Computing and Pattern Recognition}, 2009, pp. 43--48.

\bibitem{Gowersimilarity}
J.~C. Gower, ``A general coefficient of similarity and some of its
  properties,'' \emph{Biometrics}, vol.~27, no.~4, pp. 857--871, 1971.

\bibitem{ZhangKHarmonic}
B.~Zhang, ``Generalized k-harmonic means--dynamic weighting of data in
  unsupervised learning,'' in \emph{Proceedings of the 2001 SIAM International
  Conference on Data Mining}.\hskip 1em plus 0.5em minus 0.4em\relax SIAM,
  2001, pp. 1--13.

\bibitem{Zheng5586136}
Z.~Zheng, M.~Gong, J.~Ma, L.~Jiao, and Q.~Wu, ``Unsupervised evolutionary
  clustering algorithm for mixed type data,'' in \emph{IEEE Congress on
  Evolutionary Computation}, 2010, pp. 1--8.

\bibitem{Gluck}
J.~C. M.A.~Gluck, ``Information, uncertainty, and the utility of categories,''
  in \emph{Proceeding of the 7th Annual Conference of the Cognitive Science
  Society, Lawrence Erlbaum Associates, Irvine}, 1985, pp. 283--287.

\bibitem{Mirkin2001}
B.~Mirkin, ``Reinterpreting the category utility function,'' \emph{Machine
  Learning}, vol.~45, no.~2, pp. 219--228, 2001.

\bibitem{Renyi}
A.~Renyi, ``On measures of entropy and information,'' in \emph{Proceeding of
  the 4th Berkeley Symposium on Mathematics of Statistics and Probability},
  1961, pp. 547--561.

\bibitem{Langcomplement}
J.~Liang, K.~S. Chin, C.~Dang, and R.~C.~M. Yam, ``A new method for measuring
  uncertainty and fuzziness in rough set theory,'' \emph{International Journal
  of General Systems}, vol.~31, no.~4, pp. 331--342, 2002.

\bibitem{Rahman2014345}
M.~A. Rahman and M.~Z. Islam, ``A hybrid clustering technique combining a novel
  genetic algorithm with k-means,'' \emph{Knowledge-Based Systems}, vol.~71,
  pp. 345--365, 2014.

\bibitem{GABook}
M.~Mitchell, \emph{An Introduction to Genetic Algorithms (Complex Adaptive
  Systems)}.\hskip 1em plus 0.5em minus 0.4em\relax MIT Press, 1998.

\bibitem{Rahman2012}
M.~I. M.A.~Rahman, ``Crudaw: a novel fuzzy technique for clustering records
  following user defined attribute weights,'' in \emph{Data Mining and
  Analytics 2012 (AusDM 2012), Sydney, Australia, 2012}, 2012, pp. 27--42.

\bibitem{BirchZhang1997}
T.~Zhang, R.~Ramakrishnan, and M.~Livny, ``Birch: A new data clustering
  algorithm and its applications,'' \emph{Data Mining and Knowledge Discovery},
  vol.~1, no.~2, pp. 141--182, 1997.

\bibitem{Goodallsimilarity}
D.~Goodall, ``A new similarity index based on probability,'' \emph{Biometrics},
  vol.~22, pp. 882--907, 1966.

\bibitem{Conceptclustering1}
J.~Han and Y.~Fu, ``Dynamic generation and refinement of concept hierarchies
  for knowledge discovery in databases,'' in \emph{AAAIWS'94 Proceedings of the
  3rd International Conference on Knowledge Discovery and Data Mining}, 1994,
  pp. 157--168.

\bibitem{Conceptclustering2}
J.~Han, Y.~Cai, and N.~Cercone, ``Data-driven discovery of quantitative rules
  in relational databases,'' \emph{IEEE Transactions on Knowledge and Data
  Engineering}, vol.~5, no.~1, pp. 29--40, 1993.

\bibitem{ART}
G.~A. Carpenter and S.~Grossberg, ``Adaptive resonance theory,'' in
  \emph{Encyclopedia of Machine Learning}, 2010, pp. 22--35.

\bibitem{Modelbasedclustering}
V.~Melnykov and R.~Maitra, ``Finite mixture models and model-based
  clustering,'' \emph{Statistical Survey}, vol.~4, no. 80-116, 2010.

\bibitem{Dempster77maximumlikelihood}
A.~P. Dempster, N.~M. Laird, and D.~B. Rubin, ``Maximum likelihood from
  incomplete data via the em algorithm,'' \emph{Journal of the Royal
  Statistical Society, Series B}, vol.~39, no.~1, pp. 1--38, 1977.

\bibitem{Hunt12003429}
L.~Hunt and M.~Jorgensen, ``Mixture model clustering for mixed data with
  missing information,'' \emph{Computational Statistics and Data Analysis},
  vol.~41, no. 3-4, pp. 429--440, 2003.

\bibitem{Hunt2WIDM:WIDM33}
------, ``Clustering mixed data,'' \emph{Wiley Interdisciplinary Reviews: Data
  Mining and Knowledge Discovery}, vol.~1, no.~4, pp. 352--361, 2011.

\bibitem{MonteEM}
G.~McLachlan and T.~Krishnan, \emph{The EM Algorithm and Extensions}.\hskip 1em
  plus 0.5em minus 0.4em\relax WILEY, 2008.

\bibitem{McParlandmodelclustering}
D.~McParland, C.~M. Phillips, L.~Brennan, H.~M. Roche, and I.~C. Gormle,
  ``Clustering high-dimensional mixed data to uncover sub-phenotypes: joint
  analysis of phenotypic and genotypic data,'' \emph{Statistics in Medicine},
  vol.~36, no.~28, pp. 4548--4569, 2017.

\bibitem{Nelsen:2006:IC:1204326}
R.~B. Nelsen, \emph{An Introduction to Copulas (Springer Series in
  Statistics)}.\hskip 1em plus 0.5em minus 0.4em\relax Berlin, Heidelberg:
  Springer-Verlag, 2006.

\bibitem{schwarz1978}
G.~Schwarz, ``Estimating the dimension of a model,'' \emph{The Annals of
  Statistics}, vol.~6, no.~2, pp. 461--464, 03 1978.

\bibitem{ICLcriterion}
C.~Biernacki, G.~Celeux, and G.~Govaert, ``Assessing a mixture model for
  clustering with the integrated completed likelihood,'' \emph{IEEE
  Transactions on Pattern Analysis and Machine Intelligence}, vol.~22, no.~7,
  pp. 719--725, Jul 2000.

\bibitem{Honda1492403}
K.~Honda and H.~Ichihashi, ``Regularized linear fuzzy clustering and
  probabilistic pca mixture models,'' \emph{IEEE Transactions on Fuzzy
  Systems}, vol.~13, no.~4, pp. 508--516, 2005.

\bibitem{Pedrycz2002}
W.~Pedrycz, ``Collaborative fuzzy clustering,'' \emph{Pattern Recognition
  Letters}, vol.~23, no.~14, pp. 1675--1686, 2002.

\bibitem{Kohonen1982}
T.~Kohonen, ``Self-organized formation of topologically correct feature maps,''
  \emph{Biological Cybernetics}, vol.~43, no.~1, pp. 59--69, 1982.

\bibitem{Kohonen:2001:SM:558021}
T.~Kohonen, M.~R. Schroeder, and T.~S. Huang, Eds., \emph{Self-Organizing
  Maps}, 3rd~ed.\hskip 1em plus 0.5em minus 0.4em\relax Berlin, Heidelberg:
  Springer-Verlag, 2001.

\bibitem{grossberg2013adaptive}
S.~Grossberg, ``Adaptive resonance theory: How a brain learns to consciously
  attend, learn, and recognize a changing world,'' \emph{Neural Networks},
  vol.~37, pp. 1--47, 2013.

\bibitem{Yin:2002:VNM:2325783.2326851}
H.~Yin, ``Visom - a novel method for multivariate data projection and structure
  visualization,'' \emph{IEEE Transactions on Neural Networks}, vol.~13, no.~1,
  pp. 237--243, Jan. 2002.

\bibitem{Alahakoon:2000:DSM:2325773.2326577}
D.~Alahakoon, S.~K. Halgamuge, and B.~Srinivasan, ``Dynamic self-organizing
  maps with controlled growth for knowledge discovery,'' \emph{IEEE
  Transactions on Neural Networks}, vol.~11, no.~3, pp. 601--614, May 2000.

\bibitem{Furao200690}
S.~Furao and O.~Hasegawa, ``An incremental network for on-line unsupervised
  classification and topology learning,'' \emph{Neural Networks}, vol.~19,
  no.~1, pp. 90--106, 2006.

\bibitem{Carpenter1991759}
G.~A. Carpenter, S.~Grossberg, and D.~B. Rosen, ``Fuzzy art: Fast stable
  learning and categorization of analog patterns by an adaptive resonance
  system,'' \emph{Neural Networks}, vol.~4, no.~6, pp. 759 -- 771, 1991.

\bibitem{Ng:2001:SCA:2980539.2980649}
A.~Y. Ng, M.~I. Jordan, and Y.~Weiss, ``On spectral clustering: Analysis and an
  algorithm,'' in \emph{Proceedings of the 14th International Conference on
  Neural Information Processing Systems: Natural and Synthetic}, ser. NIPS'01,
  2001, pp. 849--856.

\bibitem{Calinski}
T.~Calinski and J.~Harabasz, ``A dendrite method for cluster analysis,''
  \emph{Communications in Statistics}, vol.~3, no.~1, pp. 1--27, 1974.

\bibitem{Subspaceclusteringdefinition}
L.~Parsons, E.~Haque, and H.~Liu, ``Subspace clustering for high dimensional
  data: A review,'' \emph{SIGKDD Explor. Newsl.}, vol.~6, no.~1, pp. 90--105,
  Jun. 2004.

\bibitem{Jing:2007}
L.~Jing, M.~K. Ng, and J.~Z. Huang, ``An entropy weighting k-means algorithm
  for subspace clustering of high-dimensional sparse data,'' \emph{IEEE Trans.
  on Knowl. and Data Eng.}, vol.~19, no.~8, pp. 1026--1041, 2007.

\bibitem{MDLRISSANEN1978465}
J.~Rissanen, ``Modeling by shortest data description,'' \emph{Automatica},
  vol.~14, no.~5, pp. 465 -- 471, 1978.

\bibitem{DBSCAN}
M.~Ester, H.~Kriegel, J.~Sander, and X.~Xu, ``A density-based algorithm for
  discovering clusters in large spatial databases with noise,'' in
  \emph{Proceedings of the Second International Conference on Knowledge
  Discovery and Data Mining (KDD-96)}.\hskip 1em plus 0.5em minus 0.4em\relax
  AAAI Press, 1996, pp. 226--231.

\bibitem{Du8367173}
H.~Du, W.~Fang, H.~Huang, and S.~Zeng, ``{MMDBC}: Density-based clustering
  algorithm for mixed attributes and multi-dimension data,'' in \emph{2018 IEEE
  International Conference on Big Data and Smart Computing (BigComp)}, 2018,
  pp. 549--552.

\bibitem{Fisher1987}
D.~H. Fisher, ``Knowledge acquisition via incremental conceptual clustering,''
  \emph{Machine Learning}, vol.~2, no.~2, pp. 139--172, Sep 1987.

\bibitem{GENNARI198911}
J.~H. Gennari, P.~Langley, and D.~Fisher, ``Models of incremental concept
  formation,'' \emph{Artificial Intelligence}, vol.~40, no.~1, pp. 11--61,
  1989.

\bibitem{Jardineclustering}
N.~Jardine and R.~Sibson, \emph{Mathematical Taxonomy}.\hskip 1em plus 0.5em
  minus 0.4em\relax Wiley London, 1971.

\bibitem{He2002}
Z.~He, X.~Xu, and S.~Deng, ``Squeezer: An efficient algorithm for clustering
  categorical data,'' \emph{Journal of Computer Science and Technology},
  vol.~17, no.~5, pp. 611--624, 2002.

\bibitem{michailidis1998}
G.~Michailidis and J.~de~Leeuw, ``The gifi system of descriptive multivariate
  analysis,'' \emph{Statist. Sci.}, vol.~13, no.~4, pp. 307--336, 11 1998.

\bibitem{RFC}
T.~Shi and S.~Horvath, ``Unsupervised learning with random forest predictors,''
  \emph{Journal of Computational and Graphical Statistics}, vol.~15, no.~1, pp.
  118--138, 2006.

\bibitem{threeways2}
J.~Xiong and H.~Yu, \emph{An Adaptive Three-Way Clustering Algorithm for
  Mixed-Type Data}.\hskip 1em plus 0.5em minus 0.4em\relax Springer
  International Publishing, 2018.

\bibitem{Rcitation}
\BIBentryALTinterwordspacing
R.~D.~C. Team, \emph{R: A Language and Environment for Statistical Computing},
  R Foundation for Statistical Computing, Vienna, Austria, 2008. [Online].
  Available: \url{http://www.R-project.org}
\BIBentrySTDinterwordspacing

\bibitem{Hunangimp}
G.~Szepannek, ``clustmixtype: k-prototypes clustering for mixed variable-type
  data,''
  \url{https://cran.r-project.org/web/packages/clustMixType/index.html}, 2017,
  {R} {P}ackage- Online accessed 28-January-2018.

\bibitem{Clustmdimp}
D.~McParland and I.~C. Gormley, ``clustmd: Model based clustering for mixed
  data,'' \url{https://cran.r-project.org/web/packages/clustMD/index.html},
  2017, r package- Online accessed 28-January-2018.

\bibitem{MixedR}
W.~G.~D. r, ``Clustering mixed data types in r,''
  \url{https://www.r-bloggers.com/clustering-mixed-data-types-in-r/}, 2016, r
  package- Online accessed 28-January-2018.

\bibitem{ClustofVar}
M.~Chavent and J.~S. V.~K.~Simonet, B.~Liquet, ``Clustofvar: An r package for
  the clustering of variables,'' \emph{Journal of Statistical Software},
  vol.~50, no.~13, pp. 1--16, 2012.

\bibitem{CluMixR}
M.~Hummel, D.~Edelmann, and A.~Kopp-Schneider, ``Clumix: Clustering and
  visualization of mixed-type data,''
  \url{https://cran.r-project.org/web/packages/CluMix/index.html}, 2017, {R}
  {P}ackage- Online accessed 28-January-2018.

\bibitem{kamilaR}
A.~Foss and M.~Markatou, ``kamila: Methods for clustering mixed-type data,''
  \url{https://cran.r-project.org/web/packages/kamila/index.html}, 2016, {R}
  {P}ackage- Online accessed 28-January-2018.

\bibitem{MacbarR}
M.~Marbac, C.~Biernacki, and V.~Vandewalle, ``Copules-package: Mixed data
  clustering by a mixture model of gaussian copulas,''
  \url{https://rdrr.io/rforge/MixCluster/man/Copules-package.html}, 2014, {R}
  {P}ackage- Online accessed 28-January-2018.

\bibitem{AmirAhmadDeyimpl}
A.~Alsahaf, ``mixed kmeans package,''
  \url{https://www.mathworks.com/matlabcentral/fileexchange/53489-amjams-mixed_kmeans},
  2016, {MATLAB} {P}ackage Online accessed 28-January-2018.

\bibitem{MatlabMixed}
C.~Bock, ``Mixed k-means clustering algorithm with variable discretization,''
  \url{https://www.mathworks.com/matlabcentral/fileexchange/55601-mixed-k-means-clustering-algorithm-with-variable-discretization},
  2016, {MATLAB} {P}ackage- Online accessed 28-January-2018.

\bibitem{MixtCompc++}
C.~Biernacki and V.~Kubicki, ``Mixtcomp software for full mixed data,''
  \url{https://modal.lille.inria.fr/wikimodal/doku.php?id=mixtcomp}, 2016,
  {C}++ package- Online accessed 28-January-2018.

\bibitem{Malo:2007:MDC:1420749.1420777}
E.~Malo, R.~Salas, M.~Catal\'{a}n, and P.~L\'{o}pez, ``A mixed data clustering
  algorithm to identify population patterns of cancer mortality in
  hijuelas-chile,'' in \emph{Proceedings of the 11th Conference on Artificial
  Intelligence in Medicine}, ser. AIME '07, 2007, pp. 190--194.

\bibitem{Stoliemodel}
C.~B. Storlie, S.~M. Myers, S.~K. Katusic, A.~L. Weaver, R.~G. Voigt, P.~E.
  Croarkin, R.~E. Stoeckel, and J.~D. Port, ``Clustering and variable selection
  in the presence of mixed variable types and missing data,'' \emph{Statistics
  in Medicine}, vol.~37, no.~19, pp. 2884--2899, 2018.

\bibitem{saadaoui2015dimensionally}
F.~Sa{\^a}daoui, P.~R. Bertrand, G.~Boudet, K.~Rouffiac, F.~Dutheil, and
  A.~Chamoux, ``A dimensionally reduced clustering methodology for
  heterogeneous occupational medicine data mining,'' \emph{IEEE transactions on
  nanobioscience}, vol.~14, no.~7, pp. 707--715, 2015.

\bibitem{Amirdigitalmammogram}
S.~Halawani, M.~Alhaddad, and A.~Ahmad, ``A study of digital mammograms by
  using clustering algorithms,'' \emph{Journal of Scientific and Industrial
  Research (JSIR)}, vol.~71, pp. 594--600, 2012.

\bibitem{Kuri2011Clustering}
A.~Kuri-Morales, L.~E. Cortes-Berrueco, and D.~Trejo-Banos, ``Clustering of
  heterogeneously typed data with soft computing - a case study,'' in
  \emph{Proceedings of the $10^{th}$ International Conference on Artificial
  Intelligence: Advances in Soft Computing - Volume Part II}, ser.
  MICAI'11.\hskip 1em plus 0.5em minus 0.4em\relax Berlin, Heidelberg:
  Springer-Verlag, 2011, pp. 235--248.

\bibitem{BushelPhD}
P.~R. Bushel, ``Clustering of mixed data types with application to
  toxicogenomics,'' Ph.D. dissertation, North Carolina State University, 2006.

\bibitem{Abidin2016}
Z.�Abidin, N.~Fatin�N., and R.~D. Westhead, ``Flexible model-based
  clustering of mixed binary and continuous data: application to genetic
  regulation and cancer,'' \emph{Nucleic Acids Research}, vol.~45, no.~7, p.
  e53, 2017.

\bibitem{Kassi7507121}
M.~L. Kassi, A.~Berrado, L.~Benabbou, and K.~Benabdelkader, ``Towards a new
  framework for clustering in a mixed data space: Case of gasoline service
  stations segmentation in morocco,'' in \emph{2015 IEEE/ACS 12th International
  Conference of Computer Systems and Applications (AICCSA)}, 2015, pp. 1--6.

\bibitem{morlini2010comparing}
I.~Morlini and S.~Zani, ``Comparing approaches for clustering mixed mode data:
  An application in marketing research,'' in \emph{Data Analysis and
  Classification}.\hskip 1em plus 0.5em minus 0.4em\relax Springer, 2010, pp.
  49--57.

\bibitem{cheng2009customer}
M.~Cheng, Y.~Xin, Y.~Tian, C.~Wang, and Y.~Yang, ``Customer behavior pattern
  discovering based on mixed data clustering,'' in \emph{Computational
  Intelligence and Software Engineering, 2009. CiSE 2009. International
  Conference on}.\hskip 1em plus 0.5em minus 0.4em\relax IEEE, 2009, pp. 1--4.

\bibitem{Cheng5366556}
------, ``Customer behavior pattern discovering based on mixed data
  clustering,'' in \emph{2009 International Conference on Computational
  Intelligence and Software Engineering}, Dec 2009, pp. 1--4.

\bibitem{huang1997clustering}
Z.~Huang, ``Clustering large data sets with mixed numeric and categorical
  values,'' in \emph{In The First Pacific-Asia Conference on Knowledge
  Discovery and Data Mining}, 1997.

\bibitem{Iam-On2017}
N.~Iam-On and T.~Boongoen, ``Improved student dropout prediction in thai
  university using ensemble of mixed-type data clusterings,''
  \emph{International Journal of Machine Learning and Cybernetics}, vol.~8,
  no.~2, pp. 497--510, 2017.

\bibitem{Liu2012}
N.~Liu, \emph{The Research of Intrusion Detection Based on Mixed Clustering
  Algorithm}.\hskip 1em plus 0.5em minus 0.4em\relax Springer Berlin
  Heidelberg, 2012, pp. 92--100.

\bibitem{li2008weight}
T.~Li and Y.~Chen, ``A weight entropy k-means algorithm for clustering dataset
  with mixed numeric and categorical data,'' in \emph{Fuzzy Systems and
  Knowledge Discovery, 2008. FSKD'08. Fifth International Conference on},
  vol.~1.\hskip 1em plus 0.5em minus 0.4em\relax IEEE, 2008, pp. 36--41.

\bibitem{jensen2012mining}
P.~B. Jensen, L.~J. Jensen, and S.~Brunak, ``Mining electronic health records:
  towards better research applications and clinical care,'' \emph{Nature
  Reviews Genetics}, vol.~13, no.~6, p. 395, 2012.

\bibitem{khan2017review}
S.~S. Khan and J.~Hoey, ``Review of fall detection techniques: A data
  availability perspective,'' \emph{Medical engineering \& physics}, vol.~39,
  pp. 12--22, 2017.

\bibitem{ehr}
T.~O. of~the National Coordinator~for Health Information~Technology, ``What is
  an electronic health record (ehr)?''
  \url{https://www.healthit.gov/faq/what-electronic-health-record-ehr}, 2018,
  online accessed 18-December-2018.

\bibitem{khan2017daad}
S.~S. Khan, T.~Zhu, B.~Ye, A.~Mihailidis, A.~Iaboni, K.~Newman, A.~H. Wang, and
  L.~S. Martin, ``Daad: A framework for detecting agitation and aggression in
  people living with dementia using a novel multi-modal sensor network,'' in
  \emph{Data Mining Workshops (ICDMW), 2017 IEEE International Conference
  on}.\hskip 1em plus 0.5em minus 0.4em\relax IEEE, 2017, pp. 703--710.

\bibitem{khan2018detecting}
S.~S. Khan, B.~Ye, B.~Taati, and A.~Mihailidis, ``Detecting agitation and
  aggression in people with dementia using sensors—a systematic review,''
  \emph{Alzheimer's \& Dementia}, 2018.

\bibitem{peopleanalytics}
E.~Houghton and M.~Green, ``People analytics: driving business performance with
  people data,'' CIPD, Tech. Rep., 2018.

\bibitem{du2010clustering}
K.-L. Du, ``Clustering: A neural network approach,'' \emph{Neural networks},
  vol.~23, no.~1, pp. 89--107, 2010.

\bibitem{clusteringensembles}
J.~Ghosh and A.~Acharya, ``Cluster ensembles,'' \emph{Wiley Interdisc. Rew.:
  Data Mining and Knowledge Discovery}, vol.~1, no.~4, pp. 305--315, 2011.

\bibitem{clusteringensembles2}
S.~Vega-Pons and J.~Ruiz-Shulcloper, ``A survey of clustering ensemble
  algorithms,'' \emph{International Journal of Pattern Recognition and
  Artificial Intelligence}, vol.~25, no.~03, pp. 337--372, 2011.

\bibitem{audigier2016principal}
V.~Audigier, F.~Husson, and J.~Josse, ``A principal component method to impute
  missing values for mixed data,'' \emph{Advances in Data Analysis and
  Classification}, vol.~10, no.~1, pp. 5--26, 2016.

\bibitem{aljalbout2018clustering}
E.~Aljalbout, V.~Golkov, Y.~Siddiqui, and D.~Cremers, ``Clustering with deep
  learning: Taxonomy and new methods,'' \emph{arXiv preprint arXiv:1801.07648},
  2018.

\bibitem{min2018survey}
E.~Min, X.~Guo, Q.~Liu, G.~Zhang, J.~Cui, and J.~Long, ``A survey of clustering
  with deep learning: From the perspective of network architecture,''
  \emph{IEEE Access}, vol.~6, pp. 39\,501--39\,514, 2018.

\bibitem{plant2011inconco}
C.~Plant and C.~B{\"o}hm, ``Inconco: interpretable clustering of numerical and
  categorical objects,'' in \emph{Proceedings of the $17^{th}$ ACM SIGKDD
  international conference on Knowledge discovery and data mining}.\hskip 1em
  plus 0.5em minus 0.4em\relax ACM, 2011, pp. 1127--1135.

\bibitem{zhang2016feature}
X.~Zhang, C.~Mei, D.~Chen, and J.~Li, ``Feature selection in mixed data: A
  method using a novel fuzzy rough set-based information entropy,''
  \emph{Pattern Recognition}, vol.~56, pp. 1--15, 2016.

\bibitem{tang2007feature}
W.~Tang and K.~Mao, ``Feature selection algorithm for mixed data with both
  nominal and continuous features,'' \emph{Pattern Recognition Letters},
  vol.~28, no.~5, pp. 563--571, 2007.

\bibitem{Parsons:2004}
L.~Parsons, , E.~Haque, and H.~Liu, ``Subspace clustering for high dimensional
  data: A review,'' \emph{SIGKDD Explor. Newsl.}, vol.~6, no.~1, pp. 90--105,
  2004.

\bibitem{subspacenew}
S.~Goebl, X.~He, C.~Plant, and C.~Bohm, ``Finding the optimal subspace for
  clustering,'' in \emph{2014 IEEE International Conference on Data Mining},
  2014, pp. 130--139.

\bibitem{Bansal2004}
N.~Bansal, A.~Blum, and S.~Chawla, ``Correlation clustering,'' \emph{Machine
  Learning}, vol.~56, no.~1, pp. 89--113, Jul 2004.

\bibitem{Rendon:2011:CIE:1959666.1959695}
E.~Rend\'{o}n, I.~M. Abundez, C.~Gutierrez, S.~D. Zagal, A.~Arizmendi, E.~M.
  Quiroz, and H.~E. Arzate, ``A comparison of internal and external cluster
  validation indexes,'' in \emph{Proceedings of the 2011 American Conference on
  Applied Mathematics and the $5^{th}$ WSEAS International Conference on
  Computer Engineering and Applications}, ser. AMERICAN-MATH'11/CEA'11, 2011,
  pp. 158--163.

\bibitem{Internalindices}
Y.~Liu, Z.~Li, H.~Xiong, X.~Gao, and J.~Wu, ``Understanding of internal
  clustering validation measures,'' in \emph{2010 IEEE International Conference
  on Data Mining}, Dec 2010, pp. 911--916.

\end{thebibliography}
\bibliographystyle{IEEEtran}

\begin{IEEEbiography}[{\includegraphics[width=1in,height=3.25in,keepaspectratio]{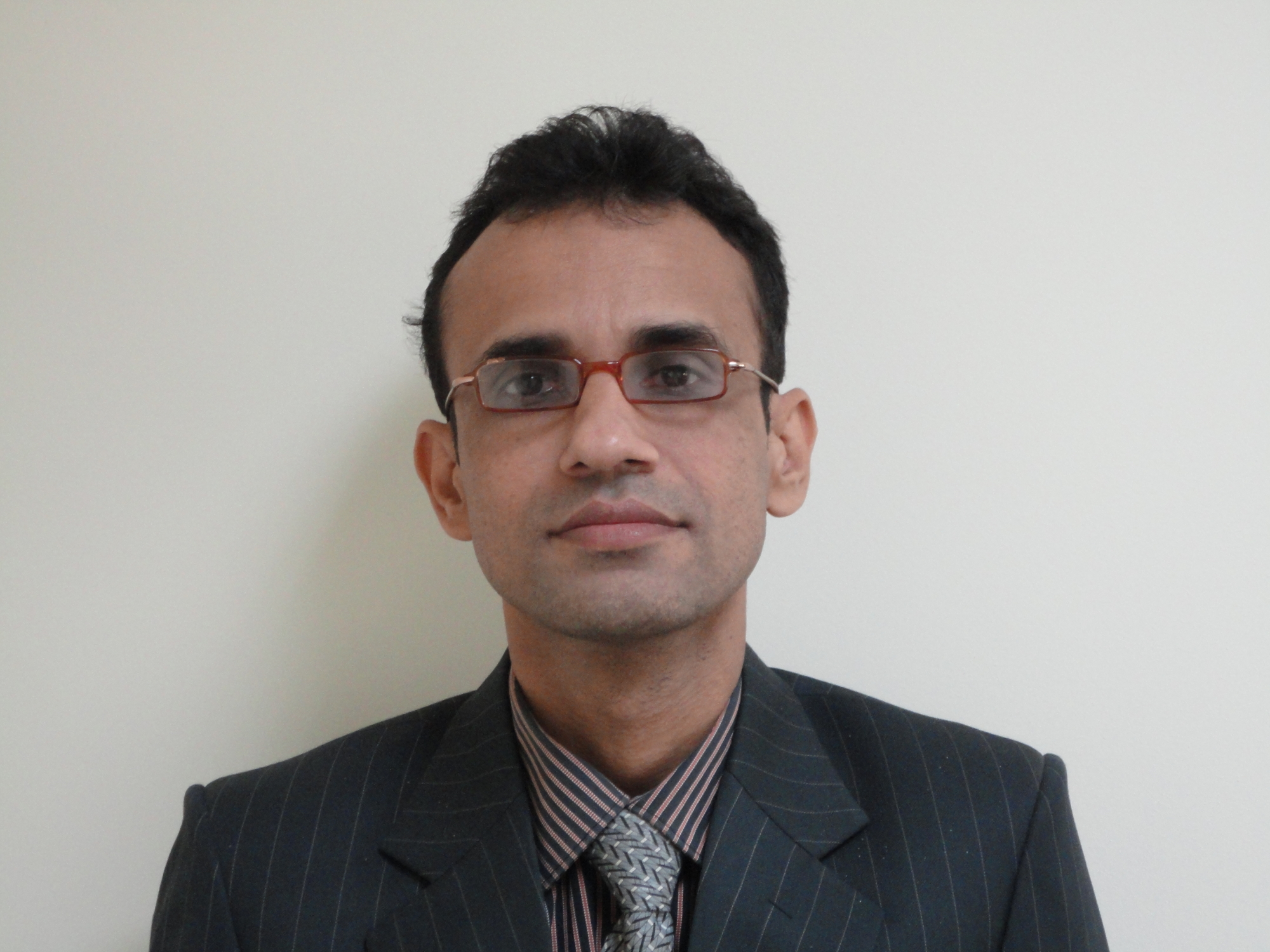}}]{Amir Ahmad} received the PhD degree in computer science from the University of Manchester, United Kingdom. He is currently working as an assistant professor in the College of Information  Technology, UAE University, Al Ain, United Arab Emirates. His research areas are machine learning, data mining, and nanotechnology.
\end{IEEEbiography}

\begin{IEEEbiography}[{\includegraphics[width=1in,height=1.25in,clip,keepaspectratio]{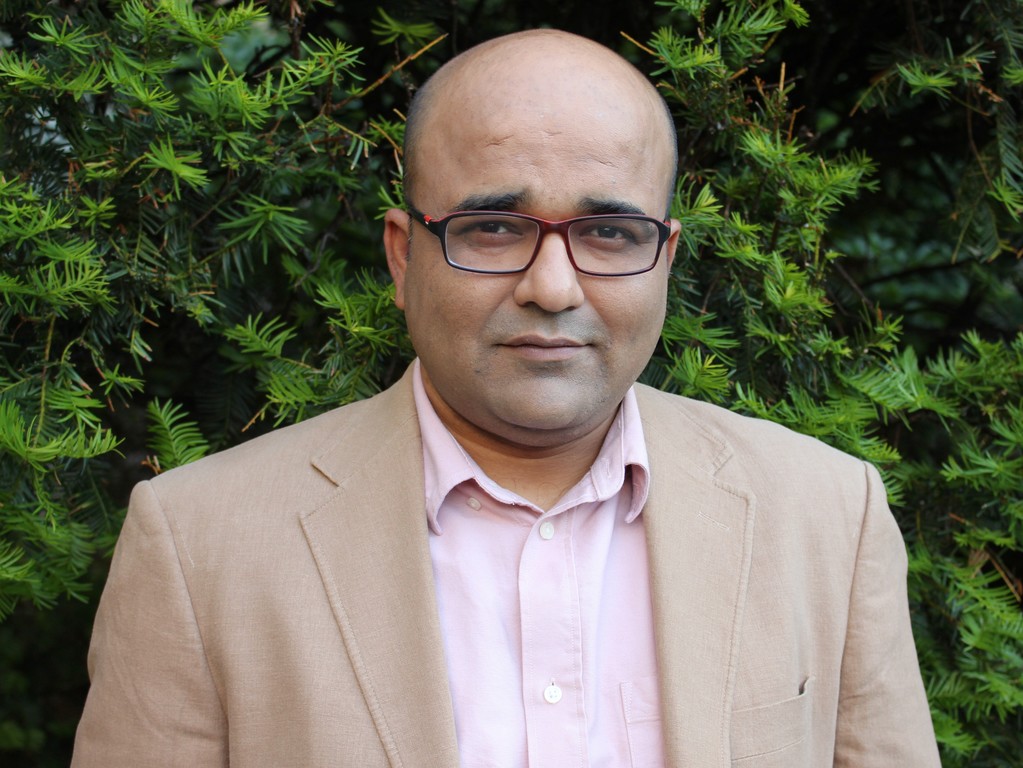}}]{SHEHROZ S. KHAN} is working as a Scientist at Toronto Rehabilitation Institute, Canada. He earned his PhD in Computer Science with specialization in Machine Learning from the University of Waterloo, Canada. He did his
Masters from National University of Ireland Galway, Republic of Ireland. Dr. Khan is also a Post-graduate Affiliate at the Vector Institute, Toronto.
His main research focus is the development of machine learning and deep learning algorithms within the realms of Aging, Rehabilitation, and Intelligent Assistive Living.
\end{IEEEbiography}

\EOD

\end{document}